\documentclass{vgtc}

\ifpdf
  \pdfoutput=1\relax
  \usepackage{graphicx}
  \DeclareGraphicsExtensions{.pdf,.png,.jpg,.jpeg}
\else
  \usepackage{graphicx}
  \DeclareGraphicsExtensions{.eps}
\fi

\graphicspath{{figures/}{pictures/}{images/}{./}} 
\usepackage{microtype}
\PassOptionsToPackage{warn}{textcomp}
\usepackage{textcomp}
\usepackage{mathptmx}
\usepackage{times}

\usepackage{cite}
\usepackage{tabu}
\usepackage{booktabs}

\usepackage{amsmath}
\usepackage{amsfonts}
\DeclareMathOperator*{\argmax}{argmax}
\DeclareMathOperator{\diag}{diag}
\DeclareMathAlphabet{\mathcal}{OMS}{cmsy}{m}{n}

\usepackage{subfig}
\usepackage[font={scriptsize,sf}]{caption}
\usepackage{enumitem}
\newlist{compactitem}{itemize}{3}
\setlist[compactitem]{label=\textbullet,leftmargin=1em, nosep}

\newcommand{\name}{FEALM}
\newcommand{\methodname}{\name{}-UMAP}

\onlineid{2112}
\vgtccategory{Research}
\vgtcpapertype{Technique}

\vgtcinsertpkg

\title{Feature Learning for Nonlinear Dimensionality Reduction toward \\Maximal Extraction of Hidden Patterns\vspace{-3pt}}

\author{
    Takanori Fujiwara\thanks{e-mail: \{takanori.fujiwara, anders.ynnerman\}@liu.se.}\\
    \scriptsize Link\"{o}ping University
    \vspace{-120pt}
\and 
    Yun-Hsin Kuo\thanks{e-mail: \{yskuo, klma\}@ucdavis.edu.}\\
    \scriptsize University of California, Davis
    \vspace{-120pt}
\and 
    Anders Ynnerman\footnotemark[1]\\
    \scriptsize Link\"{o}ping University
    \vspace{-120pt}
\and 
    Kwan-Liu Ma\footnotemark[2]\\
    \scriptsize University of California, Davis
    \vspace{-120pt}
}

\shortauthortitle{Fujiwara \MakeLowercase{\textit{et al.}}: Feature Learning for Nonlinear DR}

\abstract{
Dimensionality reduction (DR) plays a vital role in the visual analysis of high-dimensional data. One main aim of DR is to reveal hidden patterns that lie on intrinsic low-dimensional manifolds. However, DR often overlooks important patterns when the manifolds are distorted or masked by certain influential data attributes. This paper presents a feature learning framework, \textit{FEALM}, designed to generate a set of optimized data projections for nonlinear DR in order to capture important patterns in the hidden manifolds. These projections produce maximally different nearest-neighbor graphs so that resultant DR outcomes are significantly different. To achieve such a capability, we design an optimization algorithm as well as introduce a new graph dissimilarity measure, named \textit{neighbor-shape dissimilarity}. Additionally, we develop interactive visualizations to assist comparison of obtained DR results and interpretation of each DR result. We demonstrate FEALM's effectiveness through experiments and case studies using synthetic and real-world datasets.

\vspace{-4pt}

} 


\renewcommand{\manuscriptnotetxt}{}
\nocopyrightspace

\begin{document}

\firstsection{Introduction}
\maketitle

\setlength{\abovedisplayskip}{4pt}
\setlength{\belowdisplayskip}{4pt}

\newcommand{\OrgMat}{\mathbf{X}}
\newcommand{\nInsts}{n}
\newcommand{\nAttrs}{m}
\newcommand{\nLatFeats}{{m'}}
\newcommand{\nParams}{p}
\newcommand{\nEigenvals}{q}

\newcommand{\ProjMat}{\mathbf{P}}
\newcommand{\ProjMatSet}{\mathcal{P}}
\newcommand{\ReprMat}{\mathbf{Y}}

\newcommand{\DRFunc}{f_{\mathrm{DR}}}
\newcommand{\GraphFunc}{f_{\mathrm{Gr}}}
\newcommand{\Graph}{G}
\newcommand{\DRResultSet}{\mathcal{Y}}
\newcommand{\GraphResultSet}{\mathcal{G}}
\newcommand{\nResults}{r}
\newcommand{\nNonbestResults}{s}
\newcommand{\DRDissim}{d_{\mathrm{DR}}}
\newcommand{\GraphDissim}{d_{\mathrm{Gr}}}
\newcommand{\ReduceFunc}{\Phi}
\newcommand{\AttrWeights}{\mathbf{w}}
\newcommand{\UnitAttrWeights}{\mathbf{u}}
\newcommand{\OrthMan}{\mathbf{M}}
\newcommand{\LatFeatWeights}{\mathbf{v}}
\newcommand{\UnitLatFeatWeights}{\mathbf{u'}}
\newcommand{\AdjMat}{\mathbf{A}}
\newcommand{\SNNMat}{\mathbf{S}}
\newcommand{\SNNMatDiff}{\mathbf{D}}
\newcommand{\SNNDissim}{d_{\mathrm{ND}}}
\newcommand{\NetLSDDissim}{d_{\mathrm{SD}}}
\newcommand{\NSDDissim}{d_{\mathrm{NSD}}}
\newcommand{\NSDParam}{\beta}
\newcommand{\SnCDissim}{d_{\mathrm{SnC}}}
\newcommand{\LassoCoeff}{\lambda_1}
\newcommand{\RidgeCoeff}{{\lambda_2}}

\newcommand{\DRResult}[1]{Y_{#1}}
\newcommand{\GraphResult}[1]{G_{#1}}

\newcommand{\ViewNameOverview}{DR result similarity}
\newcommand{\ViewNameDRResults}{DR results}
\newcommand{\ViewNameSingleDR}{DR result examination}
\newcommand{\ViewNameInterpretation}{interpretation}
\newcommand{\ViewNameGroup}{group edition}

\newcommand{\norm}[1]{\left\lVert #1 \right\rVert}

High-dimensional data can contain a rich set of observations measured from phenomena.
Dimensionality reduction (DR) constitutes a tool for the understanding of the phenomena by visually revealing patterns in the data and facilitating human interpretation of the patterns~\cite{sacha2016visual,chatzimparmpas2020tvisne,fujiwara2020supporting}, leading to important and fundamental insights.
Among others, nonlinear DR, such as t-SNE~\cite{van2014accelerating} and UMAP~\cite{mcinnes2018umap}, is especially helpful when the patterns are hidden in nonlinear structures (or manifolds) and infeasible to be found from conventional depictions of data (e.g., with scatterplot matrices, heatmaps, and parallel coordinates~\cite{liu2016visualizing}).

However, the nonlinear DR process is sensitive to an attribute's influence on manifolds.
While nonlinear DR is commonly applied to all available attributes, it may fail to capture patterns underlying manifolds that are apparent only in a particular subset of attributes~\cite{kriegel2009clustering}. 
A similar problem also happens when manifolds are entangled by the relationships among attributes. 
Although researchers have investigated the effect of attribute selection on linear DR results~\cite{sun2021evosets}, there is a lack of studies dealing with nonlinear DR as well as the case where manifolds are entangled. 

In this work, we complement nonlinear DR methods to enable them to extract various important patterns existing in the manifolds embedded in the attributes or combinations of attributes.
We first demonstrate that nonlinear DR suffers from the aforementioned problems even when a trivial change in data, such as the inclusion of one additional attribute, is made. 
We then present a feature learning framework, \textit{\name{}}, designed to discover latent features of data, with which nonlinear DR produces significantly different results from the one using all the available attributes as they are.
These latent features can be constructed with a linear projection that is equivalent to a combination of data scaling and transformation, which are commonly used for data preprocessing~\cite{garcia2015data}. \name{}'s feature learning is performed through the maximization of the differences between data representations (e.g., nearest neighbor graphs) highly related to nonlinear DR results.
Within this framework, we design an exemplifying method for UMAP. 
We develop an algorithm utilizing the Nelder-Mead optimization method (NMM)~\cite{gao2012implementing} to find latent features to produce maximally different nearest-neighbor graphs, which are intermediate products of UMAP, and consequently generate diverse UMAP results. 
To detect the difference of the graphs, we introduce a new graph dissimilarity measure, called \textit{neighbor-shape dissimilarity} (or \textit{NSD}).
Using this method, analysts can find multiple relevant UMAP results that are difficult to find through manual preprocessing of data. 

We further develop an interactive visual interface to allow analysts to flexibly seek more patterns and gain insights from them. 
The interface depicts the similarities of DR results generated during the optimization process to notify unexplored embeddings.
Also, through brushing and linking, analysts can conveniently compare multiple DR results. 
To help review each DR result, our interface integrates an existing contrastive-learning-based interpretation method~\cite{fujiwara2020supporting}, which highlights characteristics of a group of instances through comparison with others. 

We demonstrate the effectiveness of \name{}, the exemplifying method, and the visual interface through experiments using synthetic datasets and multiple case studies on real-world datasets. 
We also conduct a performance evaluation to assess the efficiency of NSD and the optimization algorithm. 
We provide a demonstration video of the interface, detailed evaluation results, and the source code of \name{} in the supplementary material~\cite{supp}.

In summary, we consider our primary contributions to be:
\begin{compactitem}
    \item a feature learning framework, \name{}, designed to extract
    a set of latent features for nonlinear DR, each of which produces a significantly different DR result;
    \item an exemplifying method for UMAP, where we introduce an NMM-based algorithm as well as a graph dissimilarity measure, NSD;
    \item a visual interface that assists exploration of DR results and interpretation of each DR result; and
    \item designed examples that illustrate nonlinear DR's sensitiveness to trivial disturbance to intrinsic manifolds.
\end{compactitem}

\section{Related Work}

Our work supplements existing DR methods by learning appropriate features from high-dimensional data. 
Our feature learning explores various linear subspaces of the original data to generate significantly different DR results.
We provide the background and relevant works in DR methods and subspace exploration.

\subsection{Dimensionality Reduction Methods}
\label{sec:related_work_dr}

DR is widely used for visual exploration of high-dimensional data~\cite{nonato2019multidimensional,sacha2016visual}.
Many visualization-purpose DR methods aim to reveal overall data distributions (e.g., variances with principal component analysis, or PCA) or patterns (e.g., clusters in t-SNE results~\cite{van2014accelerating,fujiwara2020supporting}) in a low-dimensional space.
When only performing a linear projection~\cite{cunningham2015linear}, a DR method is categorized as linear DR. 
More precisely, it produces an embedding (or representation), $\smash{\ReprMat}$, from input data, $\smash{\OrgMat}$, with $\smash{\ReprMat = \OrgMat \ProjMat}$, where $\smash{\ReprMat {\in} \mathbb{R}^{\nInsts \times \nLatFeats}}$, $\smash{\OrgMat {\in} \mathbb{R}^{\nInsts \times \nAttrs}}$, $\smash{\ProjMat {\in} \mathbb{R}^{\nAttrs \times \nLatFeats}}$, and $\smash{\nInsts}$, $\smash{\nAttrs}$, $\smash{\nLatFeats}$ are the numbers of instances, attributes, and latent features, respectively.
For example, PCA is a well-known linear DR method.
While linear DR only captures the linear structure of $\OrgMat$, nonlinear DR can uncover the nonlinear structure~\cite{van2009dimensionality}. 
For example, t-SNE~\cite{van2014accelerating} and UMAP~\cite{mcinnes2018umap} aim to preserve local neighborhoods of each instance, which is often difficult when relying only on a linear projection.  

Despite the frequent use of the aforementioned DR methods for visual analytics~\cite{sacha2016visual}, they may fail to show important patterns when data contains noises or influential attributes to the overall data distribution.
Several DR methods have been developed to address this limitation. 
For example, discriminant analysis~\cite{cunningham2015linear,guo2007regularized}, such as linear discriminant analysis (LDA), utilizes class information to reduce noises that are irrelevant to class separation.
Contrastive learning~\cite{abid2018exploring}, such as contrastive PCA, compares two datasets to reveal patterns that are more salient in one dataset when compared to another. 
Unified linear comparative analysis~\cite{fujiwara2022ulca} flexibly incorporates the strengths of both discriminant analysis and contrastive learning. 
However, all these methods require additional information (e.g., class labels) and focus only on revealing patterns related to a well-defined analysis interest (e.g., classification). 
Thus, these methods would not be suitable when performing an early-stage exploration without complete knowledge of data and/or expected findings, which is an important task that visual analytics should support.

Other than visualization, some of DR methods can be utilized for feature selection and feature learning~\cite{zebari2020comprehensive}. 
For example, PCA and LDA are frequently used in data preprocessing for subsequent machine learning (ML) methods, such as deep neural networks or even other DR methods (e.g., t-SNE), as reducing dimensions is helpful to avoid high computational costs and the curse of dimensionality. 
As shown in a comprehensive survey by Zebari et~al.~\cite{zebari2020comprehensive}, a large portion of this type of DR targets classification tasks. 

\name{} can be seen as an unsupervised linear DR method for feature learning.
We design \name{} to preprocess data to produce significantly different nonlinear DR results from one obtained using original data as is.
\name{} does not require additional information, such as class labels; thus, it supports an early-stage exploration.

\vspace{-1pt}
\subsection{Exploration of Axis-Parallel Subspaces}
\vspace{-1pt}

There are two major types of subspaces: axis-parallel subspaces and linear subspaces.
Axis-parallel subspaces are composed of a subset of original data attributes. 
Thus, there are $2^m$ subspaces we can explore. 
On the other hand, linear subspaces consist of axes obtained through linear projections of original data. 
Although linear subspaces embrace axis-parallel subspaces, here we only describe studies on axis-parallel subspaces and the rest in \autoref{sec:related_work_linear_subspace}.

Scatterplot matrices and parallel coordinates are classic visualizations to explore axis-parallel subspaces~\cite{liu2016visualizing}.
A set of methods have been developed to improve these visualizations' scalability and usability~\cite{yuan2013dimension,wilkinson2006high}.
In response to the increase in the available number of attributes, more efforts have been devoted to comparing a large set of subspaces. 
A common approach is finding meaningful subspaces with subspace selection, visualizing each subspace's dissimilarity, and informing patterns seen in each subspace with DR~\cite{tatu2012subspace,wang2017subspace,jackle2017pattern}. 
While this approach uses DR to understand subspaces, Sun et al.~\cite{sun2021evosets} investigated subspace selection's influences on PCA and multidimensional scaling (MDS) results. 
More comprehensive descriptions of relevant works can be found in a survey by Liu et~al.~\cite{liu2016visualizing}.

Subspace clustering~\cite{parsons2004subspace,kriegel2009clustering} in the ML field shares a closely related concept with our work. 
Subspace clustering aims to find clusters within axis-parallel subspaces.
By limiting the use of data to a subset of attributes, subspace clustering can uncover clusters that are masked by irrelevant attributes.
Note that, in the visualization field, ``subspace clustering'' is often confusingly used to represent clustering \textit{of} subspaces; however, in standard ML terminology, subspace clustering performs clustering \textit{within} axis-parallel subspaces. 

Our work seeks subspaces that produce significantly different nonlinear DR results to reveal hidden patterns. 
The work by Sun et~al.~\cite{sun2021evosets} and subspace clustering methods~\cite{parsons2004subspace,kriegel2009clustering} are closely related to our work in terms of analyzing the subspace change's influence on data patterns. 
However, our work is for the use with nonlinear DR methods and provides optimization and visualization methods to find appropriate linear subspaces, which are not limited to axis-parallel subspaces.

\vspace{-1pt}
\subsection{Exploration of Linear Subspaces}
\label{sec:related_work_linear_subspace}
\vspace{-3pt}

To show high-dimensional data distributions in a selected 2D linear subspace, scatterplots and star coordinates are often used with interactive enhancements~\cite{liu2016visualizing,wang2017linear}.
We can see linear DR as a method that selects a view suitable to see some characteristics of data (e.g., variance with PCA)~\cite{gleicher2013explainers}. 
Various interactive adjustment methods for linear DR have been introduced, as summarized in surveys~\cite{nonato2019multidimensional,sacha2016visual}.

When dealing with many attributes, manually finding informative subspaces becomes almost infeasible.
Thus, researchers have designed (semi)automatic and visual recommendations. 
For example, Wang et~al.~\cite{wang2017linear} utilized LDA to suggest star coordinates that show clear cluster separations. 
Zhou et~al.~\cite{zhou2016dimension} visualized the similarities of original attributes as well as 1D subspaces to help analysts construct interesting subspaces.
Gleicher et~al.~\cite{gleicher2013explainers} utilized support-vector machines to suggest simple linear subspaces that satisfy specifications provided by analysts. 
Grassmannian Atlas~\cite{liu2016grassmannian} provides an overview of projection qualities (e.g., skewness) of all 2D subspaces sampled from the Grassmannian manifold.
Lehmann and Theisel~\cite{lehmann2015optimal} developed an optimization method to provide a set of 2D subspaces that are significantly different from each other.

Lehmann and Theisel's work~\cite{lehmann2015optimal} is the most related work as they also aimed to mine various patterns in different subspaces. 
While theirs only finds and visualizes 2D subspaces(i.e., equivalent to linear DR onto 2D planes), ours searches multidimensional subspaces and uses them as nonlinear DR inputs to uncover patterns hidden in complex data. 

\vspace{-5pt}
\section{Motivating Examples}
\label{sec:motivating_example}
\vspace{-2pt}

Using UMAP~\cite{mcinnes2018umap} as a representative nonlinear DR method, we provide concrete cases where nonlinear DR misses important  patterns. 
The datasets that we created, source code for the data generation, and comprehensive experiment results are made available in the supplementary materials~\cite{supp}. 
As shown in \autoref{fig:toy_example}-a, we first generated a dataset with three attributes, with which instances are placed around two different spherical surfaces.
This dataset has approximately 200 blue (\texttt{Sphere\,1}) and 100 orange (\texttt{Sphere\,2}) instances.
The inner sphere corresponding to \texttt{Sphere\,2} has a radius with 40\% length of the outer sphere's radius.
Also, small noises following a normal distribution are added for the placement of each instance.

When visualizing this dataset in a 2D space, linear DR can only produce a plane cut of the 3D spheres; consequently, we cannot see a clear distinction between them (see \autoref{fig:toy_example}-b). 
Any linear DR causes a similar issue when data has curved shapes, such as those in the Swiss-Rolls dataset~\cite{pedregosa2011scikit}.
On the other hand, nonlinear DR methods that aim to preserve each instance's local neighbors such as t-SNE and UMAP can separate \texttt{Spheres\,1} and \texttt{2}, as shown in \autoref{fig:toy_example}-c.

\begin{figure}[tb]
    \centering
    \includegraphics[width=0.995\linewidth]{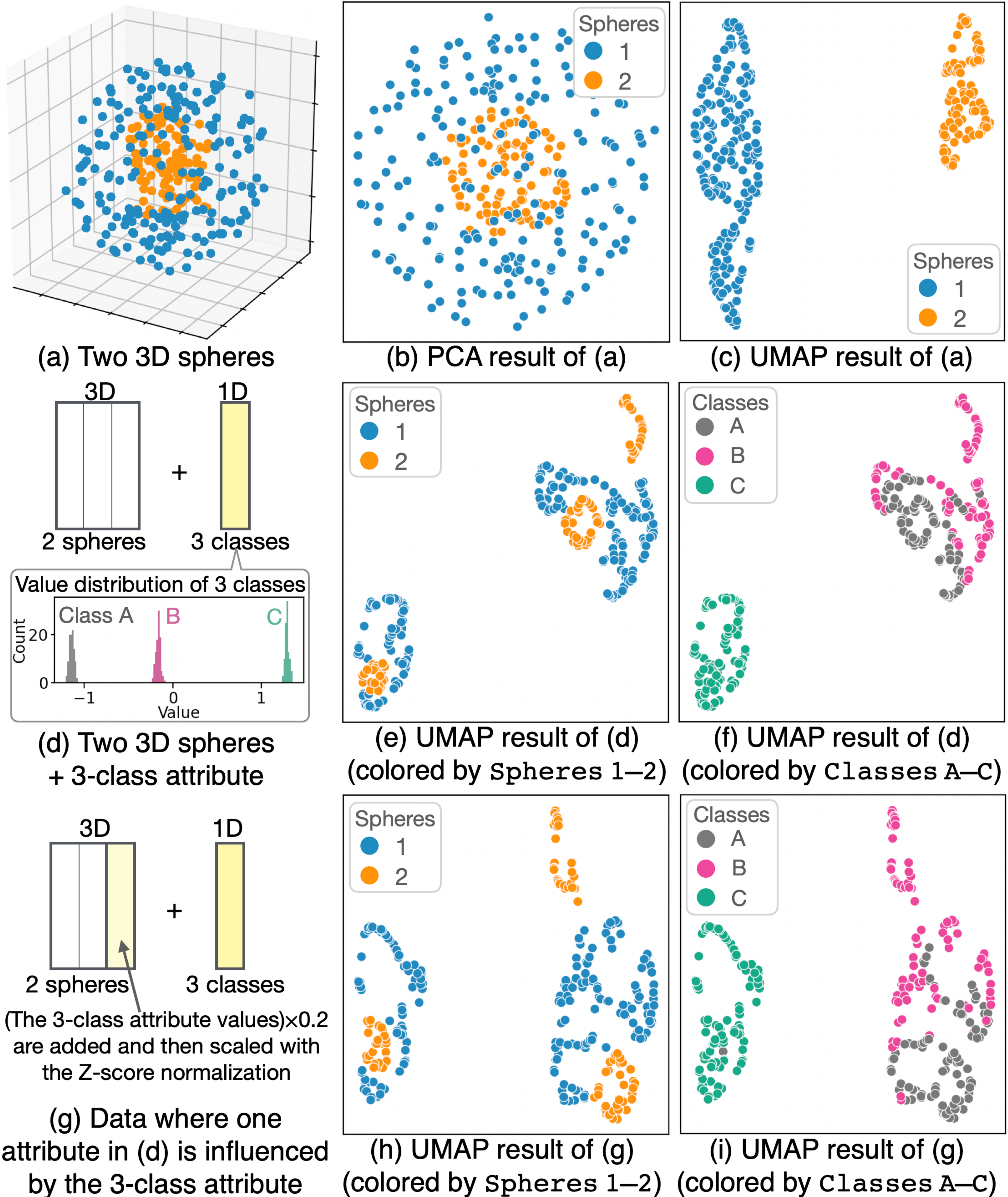}
    \caption{Three datasets (a,d,g) and corresponding DR results: PCA (b) and UMAP (c,e,f,h,i).}
    \label{fig:toy_example}
\end{figure}

However, even nonlinear DR easily fails to find important patterns when some attributes affect manifolds that contain the corresponding patterns. 
This situation can happen even with subtle changes in a dataset.
To illustrate this, as shown in \autoref{fig:toy_example}-d, we generated a new dataset by adding one attribute that has three classes (\texttt{Classes\,A}--\texttt{C}) having clearly different values. 
We shuffled the order of the 3-class attribute's instances (i.e., there are no correspondence among \texttt{Spheres\,1}--\texttt{2} and \texttt{Classes\,A}--\texttt{C}).
Also, we applied the Z-score normalization for each attribute to follow the standard preprocessing for DR~\cite{garcia2015data} and to avoid creating strong influences from the 3-class attribute. 
UMAP results for this dataset are presented in \autoref{fig:toy_example}-e, f.
We can see that UMAP does not show the separation of \texttt{Spheres\,1} and \texttt{2} anymore. 
Moreover, UMAP does not clearly distinguish \texttt{Classes\,A}--\texttt{C} either. 
Similar issues happen even when using other nonlinear DR methods, such as t-SNE.
Also, hyperparameter adjustments of UMAP, such as $k$ used for the  $k$-nearest neighbor ($k$-NN) graph construction, cannot solve this issue as the manifold itself is distorted (for details, refer to \cite{supp}).

\begin{figure}[t]
    \centering
    \includegraphics[width=0.93\linewidth,height=0.55\linewidth]{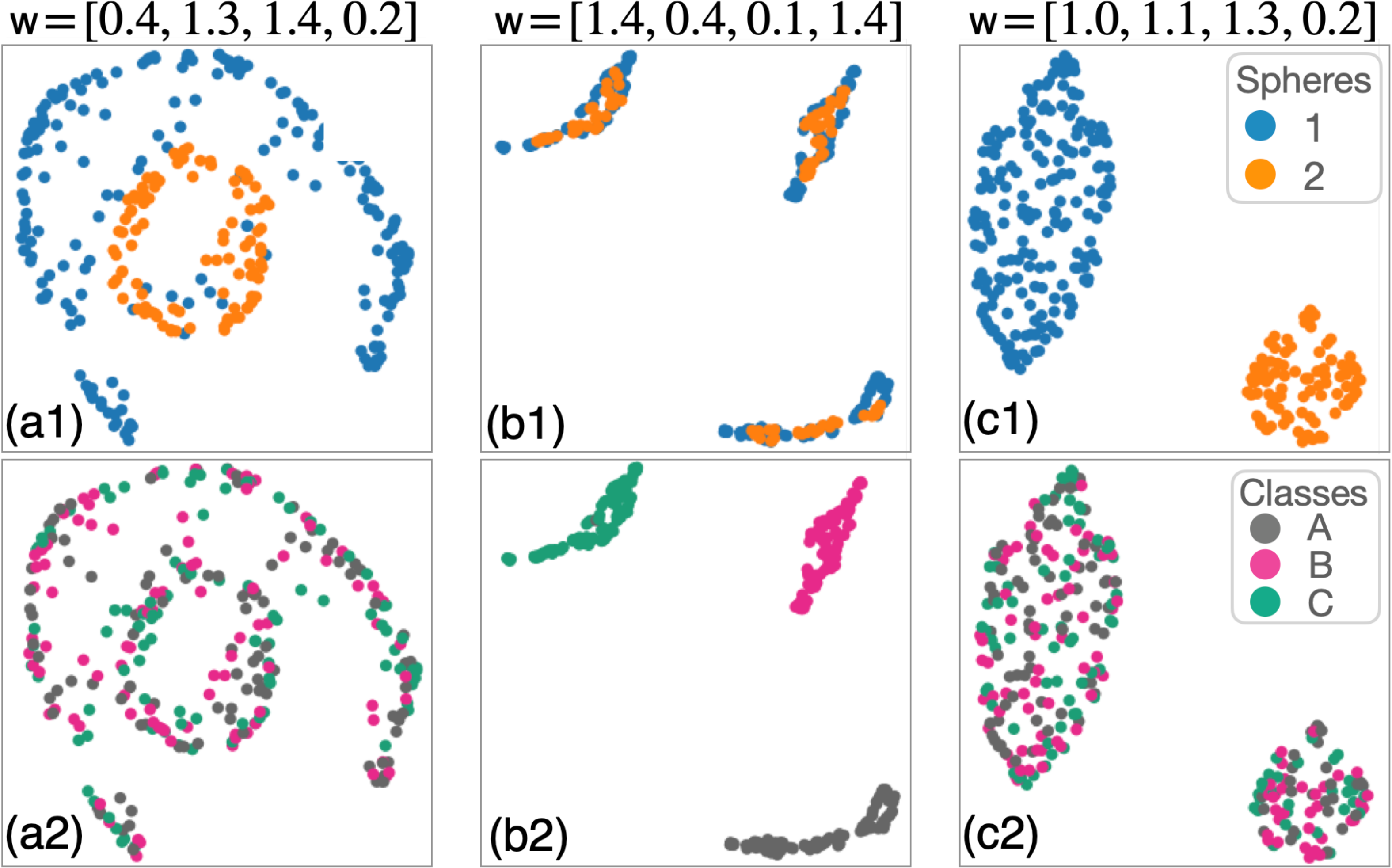}
    \caption{UMAP results of the dataset shown in \autoref{fig:toy_example}-d after using our feature learning method. $\textbf{w}$ shows the attribute weights.}
    \label{fig:toy_example_solution1}
\end{figure}

The issue seen in this dataset (\autoref{fig:toy_example}-d) can be solved by assigning larger weights to the 2-sphere attributes (or excluding the 3-class attribute), which leads to a clear separation of \texttt{Spheres\,1}--\texttt{2}.
Similarly, by assigning a large weight only to the 3-class attribute, DR can find \texttt{Classes\,A}--\texttt{C}.
In fact, our method for UMAP, which we introduce in \autoref{sec:exemplifying_method}, can reveal these two patterns by automatically adjusting the weights. 
\autoref{fig:toy_example_solution1} shows three examples suggested by our method.
The results are produced with the same UMAP's hyperparameters as those used in \autoref{fig:toy_example}-e.
In \autoref{fig:toy_example_solution1}-c1, by using a relatively small weight for the 3-class attribute (i.e., 0.2), UMAP shows clear clusters of \texttt{Spheres\,1}--\texttt{2}.
Similarly, \autoref{fig:toy_example_solution1}-b2 shows three clusters of \texttt{Classes\,A}--\texttt{C} by assigning the 3-class attribute a large weight.

\begin{figure}[t]
    \centering
    \includegraphics[width=0.93\linewidth,height=0.55\linewidth]{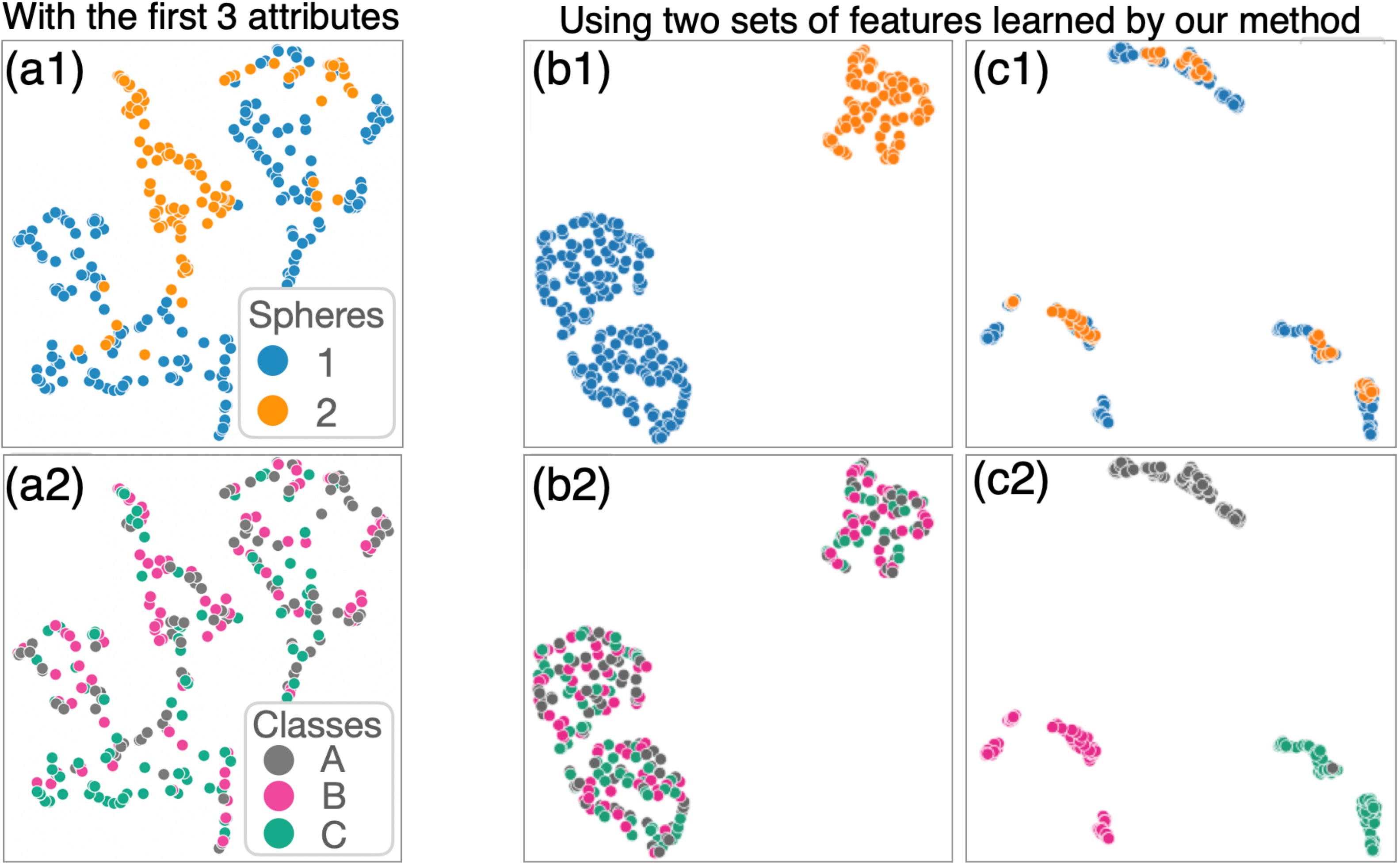}
    \caption{UMAP results of the dataset shown in \autoref{fig:toy_example}-g after selecting only the first three attributes (a) and using our feature learning (b,c).}
    \label{fig:toy_example_solution2}
\end{figure}

The above issue can be more complicated when patterns are entangled in multiple attributes' relationships. 
We create a dataset exhibiting such a case by adding 20\% portions of the 3-class attribute values into one of the 2-sphere attributes and then applying the Z-score normalization again, as described in \autoref{fig:toy_example}-g.
This type of entanglements can be found in, for example, income statistics partially influenced by age and a political opinion influenced by a voter's general ideology.
The attribute weighting or selection cannot resolve the issue in this dataset.
\autoref{fig:toy_example_solution2}-a1, a2 show UMAP results on this dataset after removing the 3-class attribute, which still do not show the separation of \texttt{Spheres\,1}--\texttt{2}.
This can be solved by learning latent features by our method.
The right two columns of \autoref{fig:toy_example_solution2} show a subset of the generated UMAP results. 
We can see b1 and c2 clearly separate \texttt{Spheres\,1}--\texttt{2} and \texttt{Classes\,A}--\texttt{C}, respectively. 

The above examples demonstrate that nonlinear DR can easily overlook important, obvious data patterns when certain attributes influence intrinsic manifolds---even a single attribute can cause this situation. 
While the examples are the case for finding distinct data groups, similar issues can happen even when finding, for example, continuous value changes on manifolds as in the Swiss-Rolls dataset.

\vspace{-3pt}
\section{\name{} Framework}
\label{sec:methodology}
\vspace{-2pt}

We introduce a feature learning framework, \textit{\name{}}, to address the stated issues in nonlinear DR. 
We name the framework \name{} because it performs \textit{FEA}ture \textit{L}earning to capture or \textit{film} patterns underlying hidden \textit{M}anifolds.
As the problem can be overly complicated based on the combinations of DR methods, their hyperparameters, and hidden manifolds we should consider, we first specify our scope.
We then describe the architecture of \name{}, where optimized projection matrices are generated with the following steps: (1) constructing a (graph) representation of data, (2) performing optimization to find a projection matrix with which a (graph) representation corresponding to projected data is maximally different from the one constructed in the first step, and (3) repeating the optimized projection matrix generation to produce a maximally different (graph) representation from those obtained so far.

\vspace{-2pt}
\subsection{Problem Scope}
\vspace{-2pt}

\name{} aims to supplement \textbf{\textit{nonlinear DR}} methods, even more specifically, for those construct a \textbf{\textit{graph-based}} data representation as their intermediate product or those generate DR results highly related to a graph-based data representation. 
This scope is reasonable and still provides enough flexibility in \name{} because many DR methods can be considered graph-based~\cite{yan2006graph,mcinnes2018umap}. 
Such methods include MDS, the Barnes-Hut t-SNE (common t-SNE implementation)~\cite{van2014accelerating}, and UMAP~\cite{mcinnes2018umap}.
For example, MDS constructs a dissimilarity matrix of instances, which can be converted to a kernel/similarity matrix---corresponding to a weighted graph where nodes and edges represent instances and their similarities, respectively.
The Barnes-Hut \mbox{t-SNE} and UMAP perform DR based on a similarity/weighted graph derived from the $k$-NN graph of instances.

\name{} searches significantly different DR results for a \textbf{\textit{given DR method}} with \textbf{\textit{given hyperparameters}}. 
\name{} is not designed to select a DR method nor hyperparameters to reveal hidden patterns.

\name{} can only find patterns underlying (nonlinear) manifolds that exist in a \textbf{\textit{linear subspace}} of the original data. 
We set this scope because of two reasons.
First, we do not want to allow unintuitive or excessive data manipulation not only to provide more interpretable data preprocessing but also to avoid leading to false patterns as much as possible.
A linear projection only allows a set of linear transformations that can be converted into a single matrix multiplication (i.e., $\ReprMat = \OrgMat \ProjMat$, as described in \autoref{sec:related_work_dr}).
Based on analysts' demands, \name{}'s linear projection can be limited to data scaling (or attribute weighting), orthogonal transformation, and a combination of both, all of which are commonly used for data preprocessing.
Second, it is computationally challenging to handle manifolds that cannot be uncovered even by applying linear projections to the data.
Such manifolds might be able to be found, for example, by utilizing neural networks; however, it requires expensive parameter tuning.  

\vspace{-2pt}
\subsection{Optimization Architecture}
\label{sec:architecture}
\vspace{-2pt}

We describe the architecture of \name{} designed with considerations of flexibility and computational efficiency. 

\noindent\textbf{Forms of the problem.} 
For input data, $\smash{\OrgMat \in \mathbb{R}^{\nInsts \times \nAttrs}}$ ($\nInsts$, $\nAttrs$: the numbers of instances and attributes), latent features can be computed with $\OrgMat \ProjMat_{i}$, where $\smash{\ProjMat_{i} \in \mathbb{R}^{\nAttrs \times \nLatFeats}}$ ($\smash{\nLatFeats}$ is the number of latent features and $\smash{\nLatFeats \leq \nAttrs}$) is a projection matrix.
With $\DRFunc$, a function that performs DR with a given method and hyperparameters, we can obtain a DR result or representation, $\ReprMat_{i}$, i.e., $\ReprMat_{i}{=}\DRFunc(\OrgMat \ProjMat_{i})$. 
We can measure a dissimilarity of two DR results, $\ReprMat_{i}$ and $\ReprMat_{j}$, with a certain function, $\DRDissim(\ReprMat_{i}, \ReprMat_{j})$ (e.g., the Frobenius norm).
This expression shows that the ``difference'' in DR results can be varied by $\DRDissim$, and it should be selected based on general analytical interests.
Let $\ReprMat_{0}$ denote a DR result using $\OrgMat$ as is (i.e., $\ReprMat_{0}{=}\DRFunc(\OrgMat)$).
Then, we can write an optimization problem to find $\ProjMat_{1}$ that generates $\ReprMat_{1}$ maximally different from $\ReprMat_{0}$ as: $\argmax_{\ProjMat_{1}} \DRDissim(\ReprMat_{1}, \ReprMat_{0})$.
When iteratively finding a new projection matrix, $\ProjMat_{i+1}$, we want to find one that generates $\ReprMat_{i+1}$ maximally different from a set of DR results already produced, $\DRResultSet_i{=}\{\ReprMat_{0}, \cdots, \ReprMat_{i}\}$. 
With some reduce function, $\ReduceFunc$ (e.g., mean), the problem of finding $\ProjMat_{i+1}$ can be written as: 
\begin{equation}
    \label{eq:maximize_dr_dissim}
    \argmax_{\ProjMat_{i+1}} \ReduceFunc \left(\DRDissim\left(\ReprMat_{i+1}, \ReprMat_{0}\right), \cdots, \DRDissim\left(\ReprMat_{i+1}, \ReprMat_{i}\right)\right).
\end{equation}

\vspace{-3pt}
Note that \autoref{eq:maximize_dr_dissim} shares some similarities with the one presented by Lehmann and Theisel~\cite{lehmann2015optimal}.  
However, their optimization is only to produce 2D linear DR results, and their case limits to $\ReprMat_{i}\, {=}\, \OrgMat \ProjMat_i$ ($\ProjMat_i \in \mathbb{R}^{\nAttrs \times 2}$). 
Also, they only use the extended version of the Procrustes distance~\cite{goodall1991procrustes} and Frobenius norm as a combination of $\DRDissim$ and $\ReduceFunc$ (for more details, refer to ~\cite{lehmann2015optimal}).
\name{} can be used for nonlinear DR and provides flexibility for each function choice.
For $\ReduceFunc$, rather than computing a norm or maximum, we recommend taking a minimum of dissimilarities (i.e., \autoref{eq:maximize_dr_dissim} maximizes the minimum of dissimilarities).
With this, we can find a DR result that is different from all the existing results and avoid a case where the optimization keeps producing the same or similar results (e.g., a case where $\ReprMat_0$ has an extremely larger dissimilarity with $\ReprMat_1$ than with other potential DR results).
We provide a graphical explanation of such a case in the supplementary materials~\cite{supp}.

While \autoref{eq:maximize_dr_dissim} is a straightforward description of our goal, directly performing this optimization for nonlinear DR methods is often difficult due to two-folded reasons.
First, $\DRFunc$ is often computationally expensive. 
For example, UMAP took 5 seconds to produce \autoref{fig:toy_example}-h from the data containing only 300 instances and 4 attributes. 
If the optimization requires many evaluations/trials, completion time can easily surpass several hours (e.g., about 1.5 hours for 1000 evaluations).
Second, popularly used nonlinear DR methods such as t-SNE and UMAP contain randomness in $\DRFunc$.
For example, random initialization of a representation (as in t-SNE) or random sampling during the optimization (as in UMAP) highly influences the final result~\cite{kobak2021initialization}.
Consequently, it becomes difficult to adjust $\ProjMat_{i}$ during the optimization---when the objective value of \autoref{eq:maximize_dr_dissim} becomes better, we do not know whether it is the improvement from changes in $\ProjMat_{i}$ or caused by the randomness in $\DRFunc$.

Thus, \name{} also introduces an optimization problem that maximizes the differences among graph representations of data.
We can use graph representations that are the same as or similar to intermediate products of given DR.
This optimization is based on a general observation: if such graph representations have maximal differences, derived DR results are also significantly different.  
Let $\GraphFunc$ denote a function that generates some graph (e.g., $k$-NN graph, similarity matrix) from an input matrix. 
Then, a graph, $\Graph_{i}$, corresponding to $\ProjMat_{i}$ can be obtained with $\Graph_{i} = \GraphFunc(\OrgMat \ProjMat_{i})$.
Also, we denote a function that measures a dissimilarity of two graphs, $\Graph_{i}$ and $\Graph_{j}$, by $\GraphDissim(\Graph_{i}, \Graph_{j})$.
With a set of already produced graphs, $\GraphResultSet_{i} = \{\Graph_{0}, \cdots, \Graph_{i}\}$, we can write a relaxed version of the optimization problem:
\vspace{-2pt}
\begin{equation}
    \label{eq:maximize_graph_dissim}
    \argmax_{\ProjMat_{i+1}} \ReduceFunc (\GraphDissim(\Graph_{i+1}, \Graph_{0}), \cdots, \GraphDissim(\Graph_{i+1}, \Graph_{i})).
    \vspace{-3pt}
\end{equation}
When developing a method within \name{}, we can choose \autoref{eq:maximize_dr_dissim} or \ref{eq:maximize_graph_dissim} based on the characteristics of a DR method, such as the computational efficiency and stability.
We can also use \autoref{eq:maximize_dr_dissim} and \ref{eq:maximize_graph_dissim} in a hybrid manner.
For example, we can generate a large number of projection matrices with \autoref{eq:maximize_graph_dissim} and then filter them with \autoref{eq:maximize_dr_dissim} to obtain refined results.

\vspace{-1pt}\noindent\textbf{Constraints on a linear projection.} 
Another important consideration of the optimization is the constraints on $\ProjMat_{i}$. 
We should decide the constraints based on data manipulation allowed for an analysis goal and the optimization difficulty for a given dataset (\autoref{sec:computational_eval} provides the detailed discussions). 
Here we list representative options: (1) no constraint; (2) allowing only data scaling; (3) allowing data scaling and orthogonal transformation. 
With any option, we can interpret how data is transformed by reviewing values in $\ProjMat_{i}$.

\vspace{-1pt}When there is (1) no constraint in a projection matrix, $\ProjMat$, \name{} most flexibly learns features. 
However, as orthogonality between each learned feature is not guaranteed, distance-related functions (e.g., $k$-NN graph construction using the Euclidean distance) might be heavily influenced by the distortion. 
Also, the optimization needs to search the best values for $\nAttrs {\times} \nLatFeats$ parameters in $\ProjMat$.

\vspace{-1pt}When (2) allowing only data scaling, $\ProjMat = \diag(\AttrWeights)$ (i.e., with $\OrgMat \ProjMat$, each column of $\OrgMat$ is multiplied by the corresponding weight in $\AttrWeights$) where $\AttrWeights$ is an $\nAttrs$-dimensional vector.
Practically, we can restrict $\smash{\AttrWeights = \sqrt{\nAttrs} \UnitAttrWeights}$ where $\UnitAttrWeights$ is a unit vector.
When $\UnitAttrWeights$ consists of uniform values, $\smash{\AttrWeights = (1 \cdots 1)^\top}$ (i.e., no scaling). 
Then, $\AttrWeights$ can be identified by searching a unit vector.
This constraint is used when generating the results in \autoref{fig:toy_example_solution1}.
As this search is only on $\nAttrs$ parameters, finding the best $\ProjMat$ is much easier than the case with no constraint. 

The last constraint (3) can be written as $\smash{\ProjMat = \diag(\AttrWeights) \OrthMan \diag(\LatFeatWeights)}$, where $\smash{\OrthMan \in \mathbb{R}^{\nAttrs \times \nLatFeats}}\!\!\!$ is an orthogonal matrix (i.e., $\smash{\OrthMan^{\!\top{}} \OrthMan = I_{\nLatFeats}}$; $\smash{I_{\nLatFeats}}$ is an $\smash{\nLatFeats {\times} \nLatFeats}$ identity matrix) and $\smash{\LatFeatWeights}$ is an $\smash{\nLatFeats}$-dimensional vector.
Here $\smash{\OrthMan}$ ensures that $\OrthMan \diag(\LatFeatWeights)$ generates orthogonal features of $\OrgMat \diag(\AttrWeights)$.
And, $\LatFeatWeights$ weights the features to control their influence on a projection. 
Similar to $\AttrWeights$, we can decompose $\LatFeatWeights$ with $\smash{\LatFeatWeights = \sqrt{\nLatFeats} \UnitLatFeatWeights}$, where $\smash{\UnitLatFeatWeights}$ is a unit vector.
A projection under this constraint resembles a combination of standard preprocessing steps (i.e., data scaling and orthogonal data transformation).
This constraint still needs to find the best values for $\smash{\nAttrs {\times} \nLatFeats}$ parameters. 

\noindent\textbf{Regularization.}
To control how strongly $\ProjMat$ can be of uniform or non-uniform values, we can \textit{optionally} apply regularization by adding a penalty term into \autoref{eq:maximize_dr_dissim} or \autoref{eq:maximize_graph_dissim}.
When applying data scaling (i.e., $\ProjMat = \diag(\AttrWeights)$), we can add an L1-norm-based penalty:  $\smash{- \LassoCoeff \norm{\AttrWeights}_1}$ ($\smash{\LassoCoeff {\in} \mathbb{R}}$).
As $\smash{\LassoCoeff}$ becomes a larger \textit{positive} value, the optimization tends to produce $\AttrWeights$ with \textit{nonuniform} values. 
On the other hand, by using large \textit{negative} $\smash{\LassoCoeff}$, $\AttrWeights$ can consist of more \textit{uniform} values (e.g., when $\smash{\LassoCoeff {=} - \infty}$, $\AttrWeights$ becomes $\smash{(1 \cdots 1)^\top}$).
Using negative $\smash{\LassoCoeff}$ is especially effective when we want to avoid generating dissimilar graphs, $\smash{\GraphResultSet_{i}}$ (or dissimilar DR results, $\smash{\DRResultSet_{i}}$), that can be derived from a selection of few attributes (e.g., when analyzing binary or ordinal data). 
Note that the L2 norm of $\AttrWeights$ is always constant (i.e., $\smash{\norm{\AttrWeights}_2} {=} \sqrt{m}$) and is not suitable for this regularization.

For the other cases (e.g., $\smash{\ProjMat = \diag(\AttrWeights) \OrthMan \diag(\LatFeatWeights)}$), similar to the above, we can control the sparsity of $\ProjMat$ with the L1-norm-based penalty: $\smash{- \LassoCoeff \mathrm{\mathbf{1}^\top_\nAttrs | \ProjMat | \mathbf{1}_\nLatFeats}}$ where $\smash{\mathbf{1}_\nAttrs {=} (1 {\cdots} 1)^\top \!\! {\in} \mathbb{R}^\nAttrs}$, $\smash{\mathbf{1}_\nLatFeats {=} (1 {\cdots} 1)^\top \!\! {\in} \mathbb{R}^\nLatFeats}$, and $\smash{\mathrm{\mathbf{1}^\top_\nAttrs | \ProjMat | \mathbf{1}_\nLatFeats}}$ is the sum of all elements of $\smash{| \ProjMat |}$.
To further regulate the difference of each column in $\ProjMat$, we can add a penalty based on the sum of each $\ProjMat$ row's L2 norm: $\smash{- \RidgeCoeff \mathbf{1}^\top_\nAttrs ((\ProjMat \circ \ProjMat) \mathbf{1}_\nLatFeats)^{1/2}}$ where $\smash{\RidgeCoeff \in \mathbb{R}}$ and $\circ$ is the Hadamard product.
With large negative $\smash{\RidgeCoeff}$, \name{} generates $\ProjMat$ in which each attribute has diverse weights across columns, and vice versa.
Note that the sum of each $\ProjMat$ row's L1 norm is equal to the sum of each $\ProjMat$ column's L1 norm; thus, the L2 norm should be used to control the differences in the columns.

\noindent\textbf{General optimization strategies.}
\autoref{eq:maximize_dr_dissim} and \ref{eq:maximize_graph_dissim} can be considered as the optimization over manifolds (or often called manifold optimization)~\cite{cunningham2015linear,townsend2016pymanopt}.
For example, for the no-constraint option (1), $\ProjMat$ can be found from the Euclidean manifold. 
A solution under the constraint (2) is derived by finding $\UnitAttrWeights$ from a unit sphere manifold.
Lastly, for the constraint (3), we can find $\UnitAttrWeights$, $\UnitLatFeatWeights$ from unit sphere manifolds and $\OrthMan$ from the Grassmann manifold~\cite{townsend2016pymanopt} that is a manifold of $\nLatFeats$-dimensional subspaces of $\nAttrs$-dimensional space.
To perform manifold optimization, we can utilize existing libraries, such as Pymanopt~\cite{townsend2016pymanopt}.
These libraries can help us, for example, generate parameters on a specified manifold. 

To solve the optimization, when all functions involved in \autoref{eq:maximize_dr_dissim} (or \autoref{eq:maximize_graph_dissim}) are differentiable, we can utilize automatic differentiation together with a solver for differentiable functions (e.g., gradient descent) through existing libraries~\cite{townsend2016pymanopt}.
When some functions are not differentiable (e.g., $k$-NN graph construction), we can use a derivative-free solver, such as the NMM.

In addition to the problem, constraints, and solver, we need to select or design $\GraphFunc$, $\GraphDissim$, and/or $\DRDissim$ based on a DR method.

\vspace{-3pt}
\section{Exemplifying Method}
\label{sec:exemplifying_method}
\vspace{-2pt}

We design an exemplifying method for UMAP, using \name{}. 
In the rest of the paper, we denote this method \textit{\methodname{}}.
We chose UMAP because it is computationally rather efficient (e.g., when compared with t-SNE)~\cite{mcinnes2018umap} and frequently used for visualization in various applications~\cite{dorrity2020dimensionality,fujiwara2021visual,kobak2021initialization}.
The specific designs for UMAP can also be generalized and easily adapted to other DR methods. 
For example, we expect that a method for t-SNE can be developed based on \methodname{} with minor adjustments in $\GraphFunc$.

\vspace{-2pt}
\subsection{Graph Generation Function}
\vspace{-2pt}

UMAP processes data in two steps: graph construction and graph layout. 
Through iterative optimization, the graph layout process performs the placement of instances (often in 2D) based on a constructed graph. 
This iterative optimization involves random sampling and expensive computations. 
Thus, we design a method using \autoref{eq:maximize_graph_dissim}, which requires $\GraphFunc$ and $\GraphDissim$.
During the graph construction process, UMAP computes the instance dissimilarities (by default, using the Euclidean distance) and then produces a $k$-NN graph based on the dissimilarities.
Afterward, UMAP constructs a fuzzy graph, which is a weighted graph where the dissimilarities are converted to the fuzzy topological representation (refer to \cite{mcinnes2018umap} for details).
From this fact, $\GraphFunc$ can be the generation of a $k$-NN or fuzzy graph. 
While a fuzzy graph contains richer information of instances' relationships, many of the state-of-the-art graph dissimilarity measures, including those we utilize to design our measure, are only available for unweighted graphs~\cite{mccabe2021netrd}.
Therefore, we use $\GraphFunc$ that generates a \textbf{\textit{$\textit{\textbf{k}}$-NN graph}}; however, we can replace this with a fuzzy graph once $\GraphDissim$ suitable for weighted graphs is developed.

\vspace{-2pt}
\subsection{Graph Dissimilarity Measure}
\label{sec:graph_dissim_measure}
\vspace{-2pt}

We can select a dissimilarity measure for unweighted graphs based on analysis interest.
When using nonlinear DR for visualization, however, we usually want to reveal patterns that are visually apparent and related to the instances' neighborhood relationships (i.e., \textit{shape} and \textit{neighbors}), such as clusters and outliers~\cite{nonato2019multidimensional}.
Another critical consideration is computational efficiency as graph comparison itself is often expensive.
But, it is difficult to judge only based on theoretical time complexity because of their detailed implementation differences (e.g., requiring only fast matrix operations or slow iterative loops). 
Based on our experiments (see Sec. C.2 in our supplementary materials~\cite{supp}), we identify that NetLSD~\cite{tsitsulin2018netlsd} can better capture differences of graph shapes with a greatly shorter runtime than many other measures available in a library of graph dissimilarities~\cite{mccabe2021netrd}.
With the eigenvalue-based approximation~\cite{tsitsulin2018netlsd}, NetLSD's time complexity is $\smash{\mathcal{O}(\nEigenvals k \nInsts + \nEigenvals^2 \nInsts)}$, where $\nEigenvals$ is the total number of the top and bottom eigenvalues to take for the approximation and $k$ is the number of neighbors set to generate an $k$-NN graph.
NetLSD does not consider the neighbor dissimilarity (i.e., the difference of neighborhood relationships); thus, for $\GraphDissim$, we introduce a new measure, \textbf{\textit{NSD}}, to capture both neighbor and shape dissimilarities. 

\vspace{-3pt}
\subsubsection{Neighbor-Shape Dissimilarity (NSD)}
\label{sec:nsd}
\vspace{-2pt}

\begin{figure}[t]
    \centering
    \includegraphics[width=0.95\linewidth,height=0.5\linewidth]{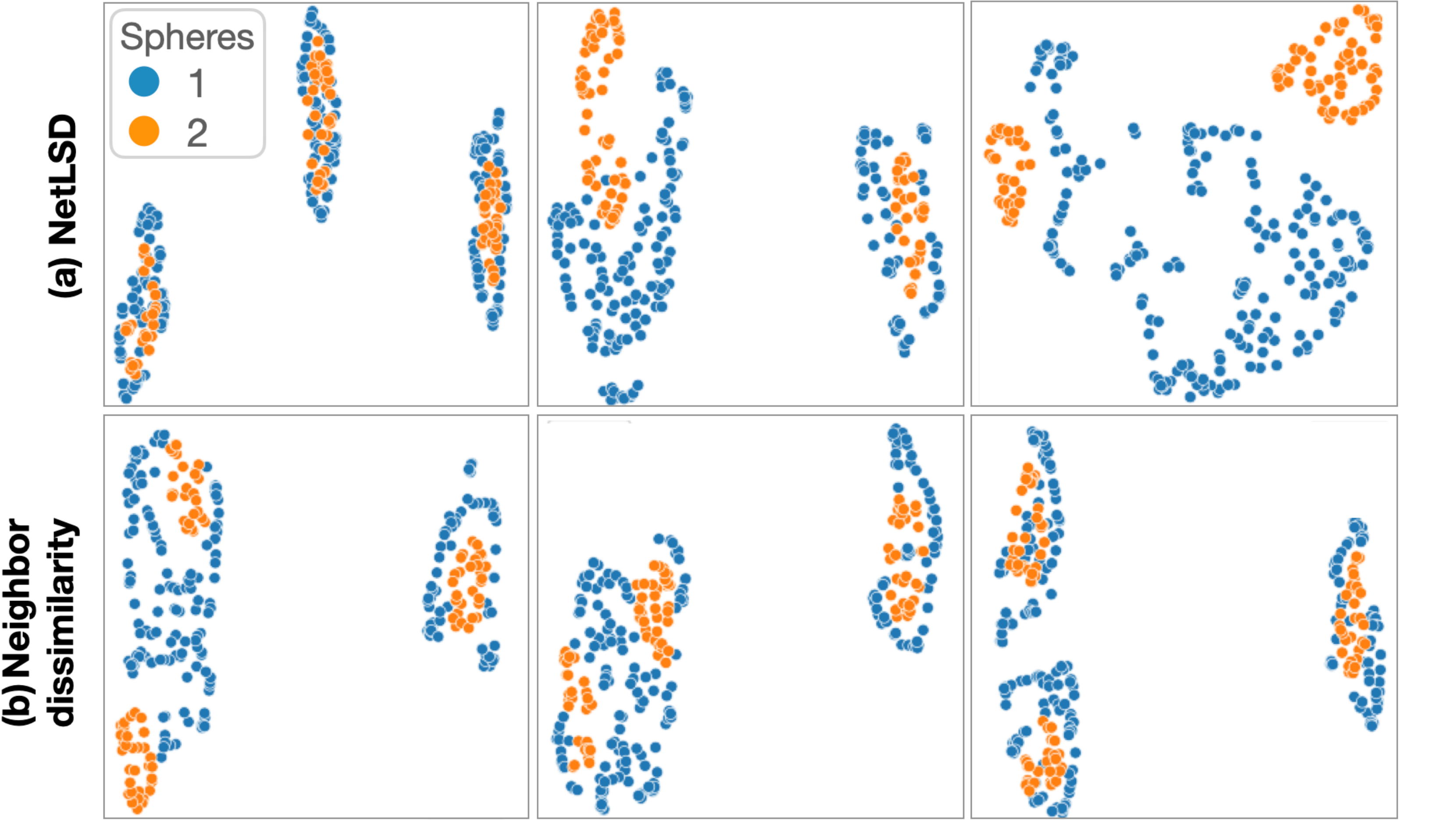}
    \caption{UMAP results of the dataset shown in \autoref{fig:toy_example}-d, using the same settings with \autoref{fig:toy_example_solution1} except for $\GraphDissim$. While \autoref{fig:toy_example_solution1} is generated with NSD, here we use NetLSD (a) and the neighbor dissimilarity (b).}
    \label{fig:netlsd_nd}
\end{figure}

We design a neighbor dissimilarity measure, \textit{ND}, and combine it with NetLSD to introduce \textit{NSD}.
\autoref{fig:toy_example_solution1} and \autoref{fig:netlsd_nd} demonstrate how these measures affect DR results when applying \methodname{} to the same dataset.
For example, as seen in \autoref{fig:netlsd_nd}-b, ND only considers the changes of $k$-neighbors around each instance; as a result, they tend to form similar shapes, where the orders of adjacent nodes are likely different.
On the other hand, the results with NetLSD (\autoref{fig:netlsd_nd}-a) show different shapes; however, they might not involve many neighborhood changes. 
Although we need more investigations to precisely conclude these tendencies, the algorithms used for NetLSD and ND only consider the shape and neighbor differences, respectively.
When using NSD (\autoref{fig:toy_example_solution1}), we can find patterns related to both types of changes. 

For the neighbor dissimilarity, one option is to utilize the steadiness and cohesiveness (SnC)~\cite{jeon2022measuring}, which are developed to assess the DR quality by measuring the changes of the neighborhood relationships in the original data and a DR result.
While SnC is adaptable to graph comparison, it involves random-walk-based sampling and expensive clustering steps.
Thus, similar to the reasons why using \autoref{eq:maximize_graph_dissim} instead of \autoref{eq:maximize_dr_dissim}, SnC is not suitable for use in the optimization.
Inspired by SnC, we design ND based on shared-nearest neighbor (SNN) similarity~\cite{ertoz2003finding}. 

SNN similarity measures how much of neighbors are shared in each pair of instances in a graph. 
Let $\AdjMat$ denote a directed adjacency matrix containing the information of each instance's $k$-NNs.
Then, all instance pairs' SNN similarities, $\SNNMat$, can be computed with $\smash{\SNNMat = \AdjMat \AdjMat^\top / k}$.
Given two graphs, $\Graph_i$ and $\Graph_j$, we can obtain the difference of each instance's SNN similarity with $\smash{\SNNMatDiff_{i,j} = \SNNMat_i - \SNNMat_j}$.
Let $\smash{\SNNMatDiff_{i,j}^+}$ and $\smash{\SNNMatDiff_{i,j}^-}$ denote matrices only taking positive and negative values of $\smash{\SNNMatDiff_{i,j}}$, respectively. 
Then, $\smash{\SNNMatDiff_{i,j}^+}$ and $\smash{\SNNMatDiff_{i,j}^-}$ capture the increase and decrease of SNNs for each instance in $\Graph_i$ when compared to $\Graph_j$.
We can compute the total increase and decrease with the Frobenius norm, i.e., $\smash{\lVert \SNNMatDiff_{\scriptscriptstyle{i,j}}^+ \rVert_{F}}$ and $\smash{\lVert \SNNMatDiff_{\scriptscriptstyle{i,j}}^- \rVert_{F}}$.
Lastly, we reduce them to one value by taking the maximum. 
That is, ND of $\smash{\Graph_i}$ and $\smash{\Graph_j}$ is defined as:
\begin{equation}
    \label{eq:snn_dissim}
    \SNNDissim(\Graph_i, \Graph_j) = \max(\lVert \SNNMatDiff_{i,j}^+ \rVert_{F}, \lVert \SNNMatDiff_{i,j}^- \rVert_{F}).
\end{equation}
Unlike SnC, ND involves only simple matrix computations while keeping a similar strength to SnC to capture the neighbor dissimilarity. 
We compare SnC and ND in \autoref{sec:computational_eval}.

Let $\NetLSDDissim$ ($\NetLSDDissim \geq 0$) denote the shape dissimilarity measure using NetLSD. 
Since NetLSD is only for undirected graphs, we use undirected $k$-NN graphs as NetLSD's inputs (i.e., $\smash{\AdjMat + \AdjMat^\top - \AdjMat \circ \AdjMat^\top}$ instead of $\AdjMat$). 
Also, by default, we set $\nEigenvals = 50$ for the approximation.
Then, we define the dissimilarity measured by NSD as:
\vspace{-3pt}
\begin{equation}
    \label{eq:nsd}
    \NSDDissim(\Graph_i, \Graph_j) = \SNNDissim(\Graph_i, \Graph_j)^\NSDParam \cdot \log\left(1 + \NetLSDDissim(\Graph_i, \Graph_j)\right)
\end{equation}
where $\smash{\NSDDissim(\Graph_i, \Graph_j) \geq 0}$ and $\smash{\NSDParam}$ ($\smash{0 \leq \NSDParam \leq \infty}$) is a hyperparameter that controls how strongly NSD focuses on the neighbor dissimilarity vs.\ the shape dissimilarity. 
When $\NSDParam = 0$, NSD is equivalent to using NetLSD. 
As $\NSDParam$ increases, ND becomes more influential on NSD. 
Based on our experiment, we set $\NSDParam = 1$ by default. 
$\NSDParam$\,can\,be adjusted based on the patterns we look for.
Since NetLSD involves an exponential function when computing the dissimilarity (refer\,to \cite{tsitsulin2018netlsd}), we take a logarithm of $1 + \NetLSDDissim$ ($1$ is added to avoid taking a logarithm of $0$) to avoid excessive influence from the shape difference.

\subsection{Optimization}
\label{sec:optimization}

\methodname{} optimizes \autoref{eq:maximize_graph_dissim} while using the $k$-NN graph construction as $\smash{\GraphFunc}$ and $\smash{\NSDDissim}$ as $\smash{\GraphDissim}$.
As recommended, we use $\smash{\ReduceFunc}$ to take a minimum of the dissimilarities.
For the constraints of a linear projection, \methodname{} supports all the three representative options described in \autoref{sec:architecture}.
Since the $k$-NN graph construction is a non-differentiable function, we develop an \textit{\textbf{NMM-based derivative-free solver}}.
We provide pseudocode in the supplementary material~\cite{supp}.

The optimization by the ordinary NMM begins with initial $(\nParams + 1)$ solutions in a $\nParams$-dimensional space, where $\nParams$ is the number of parameters (i.e., when only allowing data scaling, $\nParams = \nAttrs$; for the other cases, $\smash{\nParams = \nAttrs \times \nLatFeats}$). 
The initial solutions are typically generated at random.
Then, based on the evaluation of each solution's objective value, the NMM iteratively moves each solution toward a direction along which a better solution can be likely found while gradually shrinking a searching space.
Then, after the user-indicated number of evaluations or the convergence, the NMM returns the best solution so far as a final result.
When compared with other derivative-free solvers such as the particle-swarm optimization~\cite{kennedy1995particle}, the NMM does not involve many evaluations of the objective function, and efficiently finds a reasonable solution~\cite{zahara2009hybrid}.
By employing the NMM, we can avoid an excessive number of graph dissimilarity calculations that are part of the objective function.

However, based on the initial solutions, the ordinary NMM easily falls into the local minimum.
To mitigate this issue, similar to other hybrid approaches of global and local optimization solvers~\cite{zahara2009hybrid}, we incorporate random search optimization into the NMM. 
Specifically, instead of $(\nParams + 1)$ initial solutions, our solver generates a large number of random solutions (by default, $\smash{(10 \nParams + 1)}$ solutions). 
Then, the solver selects the $(\nParams + 1)$ best solutions and applies the NMM to them to find the refined solution. 
For this refining step, to achieve faster convergence than the ordinary NMM for a case with large $\nParams$ (e.g., $\nParams > 5$), we employ the adaptive NMM introduced by Gao and Han~\cite{gao2012implementing}.
Even with this adaptive method, the NMM is usually suitable for the optimization involving a considerably small number of parameters (e.g., $\nParams < 30$). 
Thus, when $\OrgMat$ has extremely large $\nAttrs$ (e.g., $\nAttrs = 100$), we recommend preprocessing $\OrgMat$ with, for example, PCA or clustering to generate compressed attributes.
This type of approaches is often recommended for a complex optimization (e.g., t-SNE~\cite{maaten2008visualizing} often employs PCA when $\nAttrs > 30$).

Random initialization of solutions and restriction of their movement on a specified manifold (e.g., the Grassmann manifold) can be easily achieved by utilizing the manifold optimization libraries~\cite{townsend2016pymanopt}.
By repeating the above optimization, for example, till the evaluation result of \autoref{eq:maximize_graph_dissim} converges (refer to \cite{supp}), we can obtain a set of projection matrices, $\smash{\ProjMatSet = \{\ProjMat_0, \cdots, \ProjMat_\nResults\}}$ where $\nResults$ is the number repeats.
When $\nResults$ is large (e.g., $\nResults = 100$), we can perform spectral clustering~\cite{ng2001spectral} on $\ProjMatSet$ to recommend a small number of projections (e.g., 10 projections) that produce significantly different DR results.

\subsection{Implementation Details and Complexity Analysis}
\label{sec:implementation_and_complexity}

\noindent\textbf{Implementation.}
\name{} and \methodname{} are implemented with Python and libraries for matrix computations and optimizations: NumPy/SciPy, Scikit-learn~\cite{pedregosa2011scikit}, and Pymanopt~\cite{townsend2016pymanopt}.
While many graph dissimilarities, including NetLSD, are available in netrd~\cite{mccabe2021netrd}, we use our implementation, which fully utilizes matrix computations to achieve faster calculation (e.g., our implementation of NetLSD is approximately 20 times faster~\cite{supp}).
Moreover, Pathos is utilized to use multiprocessing for the NMM-based solver. 

\noindent\textbf{Time complexity analysis.}
The $k$-NN graph construction used for $\GraphFunc$ has $\smash{\mathcal{O}(\nInsts \log(\nInsts) \nAttrs}$) when using a ball-tree method~\cite{pedregosa2011scikit}.
NSD is composed of NetLSD ($\smash{\mathcal{O}(\nEigenvals k \nInsts + \nEigenvals^2 \nInsts)}$) and ND ($\smash{\mathcal{O}(k \nInsts^2)}$), where $\nEigenvals$ is the number of eigenvalues (see \autoref{sec:graph_dissim_measure}); because $\nEigenvals \leq \nInsts$, NSD has $\smash{\mathcal{O}( \nEigenvals^2 \nInsts + k \nInsts^2)}$. 
Thus, when computing the best solution with the NMM, the cost calculation for each solution takes $\smash{\mathcal{O}(\nResults \nInsts (\nEigenvals^2 + k \nInsts))}$, where $\nResults$ is the number of produced graphs so far.

\section{Computational Evaluations}
\label{sec:computational_eval}

We evaluate the performance of computations related to \methodname{} as well as the design of ND by comparing it with SnC~\cite{jeon2022measuring}. As an experimental platform, we used the MacBook Pro (16-inch, 2019) with 2.3 GHz 8-Core Intel Core i9 and 64 GB 2,667 MHz DDR4. 
We prepared datasets with the data generation code provided in \cite{fujiwara2022ulca}.
From the 20 Newsgroups dataset~\cite{uci_mlr}, their code can generate data with various numbers of instances (documents) and attributes (topics) by utilizing the latent Dirichlet allocation. 
All source code used for the evaluations is available online~\cite{supp}.

\noindent\textbf{Comparison of ND and SnC.}
We have introduced ND as a faster, more stable alternative to SnC. 
Here we validate that ND and SnC similarly capture the neighbor changes. 
Analogous to ND's $\smash{\lVert \SNNMatDiff_{i,j}^+ \rVert_{F}}$ and $\smash{\lVert \SNNMatDiff_{i,j}^- \rVert_{F}}$, SnC produces two distinct values, the steadiness and cohesiveness.
We define the SnC-based dissimilarity measure as $\SnCDissim = 1 - \min(steadiness, cohesiveness)$. 
Note that steadiness and cohesiveness take a range of 0--1; the larger, the fewer changes.
For this experiment, we set $\nInsts {=} 200, \nAttrs {=} 10$, and $k {=} 15$ and randomly generated 500 different projection matrices with the size of $10 {\times} 5$ and graphs corresponding to the projection matrices (with $\GraphFunc(\OrgMat \ProjMat)$).

\begin{figure}[tb]
    \centering
    \includegraphics[width=\linewidth]{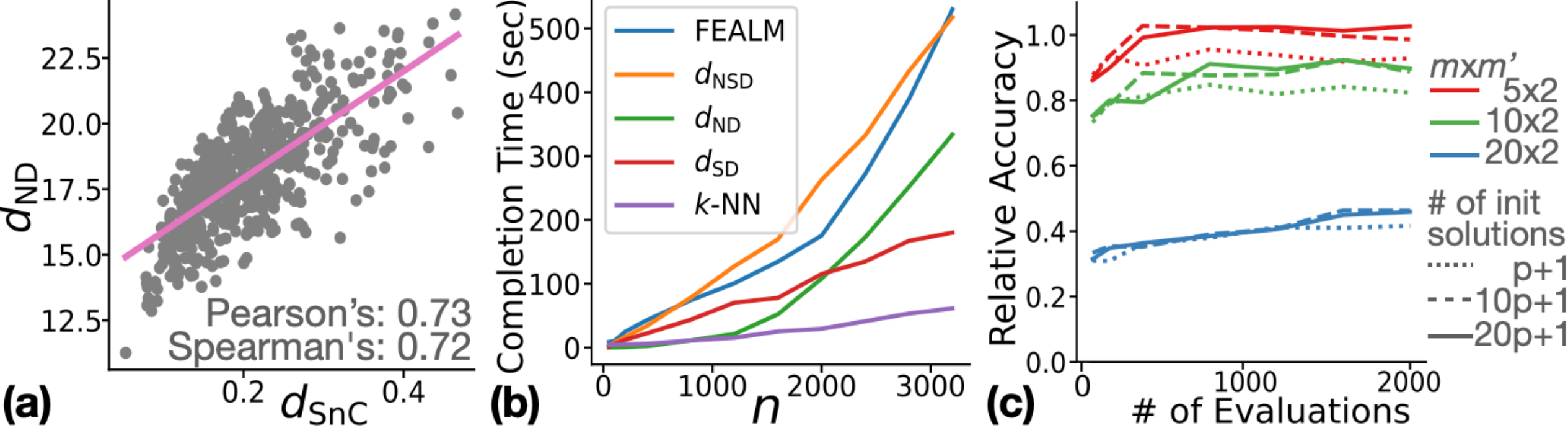}
    \caption{Computational evaluation results: (a) relationships between SnC-based dissimilarity and ND; (b) completion time for 1000 executions; (c) relative accuracy of solutions.}
     \label{fig:comp_eval}
\end{figure}

\autoref{fig:comp_eval}-a shows $\SNNDissim$ and $\SnCDissim$ of a graph corresponding to the original data and each of the 500 generated graphs. 
As SnC contains the randomness and $\SnCDissim$ can be inconsistent, we took the mean of 50 executions in \autoref{fig:comp_eval}-a. 
The mean $\SnCDissim$ of $50 {\times} 500$ results was 0.20 and the mean of 500 standard deviations was 0.02 (i.e., 10\% of the mean). 
\autoref{fig:comp_eval}-a presents strong correlations between $\SNNDissim$ and $\SnCDissim$ with Pearson's and Spearman's correlation coefficients of 0.73 and 0.72, respectively. 
Thus, similar to SnC, ND captures the neighbor changes, while ND has no randomness and a significantly smaller computational cost, as discussed in the performance evaluation below.
ND's strengths enable us to provide computationally efficient, stable NSD.

\noindent\textbf{Performance of $\DRFunc$, $\GraphFunc$, $\GraphDissim$, and the optimization.}
We evaluate the efficiency of functions related to \autoref{eq:maximize_dr_dissim} and \ref{eq:maximize_graph_dissim}, specifically, UMAP ($\DRFunc$), $k$-NN graph construction ($\GraphFunc$), ND ($\SNNDissim$), NetLSD ($\NetLSDDissim$), NSD ($\NSDDissim$), and SnC.
The number of instances, $\nInsts$, dominates these functions' complexities. 
Thus, we ran the functions with different $\nInsts$ ($\nInsts {=} 50, \cdots, 3200$) but fixed $k$, $m$, and $\nEigenvals$: $k {=} 15$ (UMAP's default), $\nAttrs {=} 10$, and $\nEigenvals {=} 50$ (NSD's default). 
As we expect many executions for the NMM, we measured the completion time of 1000 executions.

As expected, UMAP and SnC spent much longer completion time than others: e.g., 1616 and 527 seconds, respectively, for 1000 executions when $\nInsts {=} 50$. 
Therefore, these functions are not suitable to be used in optimizations that require many evaluations or deal with a larger $\nInsts$---this is the reason why we have designed \autoref{eq:maximize_graph_dissim} and ND.
Other functions' completion times are shown in \autoref{fig:comp_eval}-b.
We observe that, as $\nInsts$ increases, $\SNNDissim$ requires more computations, and dominates the completion time of $\NSDDissim$.
However, 1000 executions of $\NSDDissim$ still can be completed within 520 seconds when $\nInsts {=} 3200$.

We next evaluate the performance of \methodname{} as a whole. 
In addition to the same settings above, we set $\nLatFeats {=} 2$, no constraint on a projection matrix, and 1000 as the number of objective function evaluations. 
We then generated a single UMAP result through the optimization (i.e., $\nResults {=} 1$).
From the result shown in \autoref{fig:comp_eval}-b, we can see the completion time of \methodname{} generally follows $\NSDDissim$.
Even though \methodname{} involves more computations (e.g., $k$-NN graph construction and the solution update for the NMM), \methodname{} tends to show faster completion than $\NSDDissim$. 
This is probably because our implementation of the NMM partially employs parallel computations, as described in \autoref{sec:implementation_and_complexity}. 

\begin{figure*}[t]
    \centering
    \includegraphics[width=\linewidth,height=0.44\linewidth]{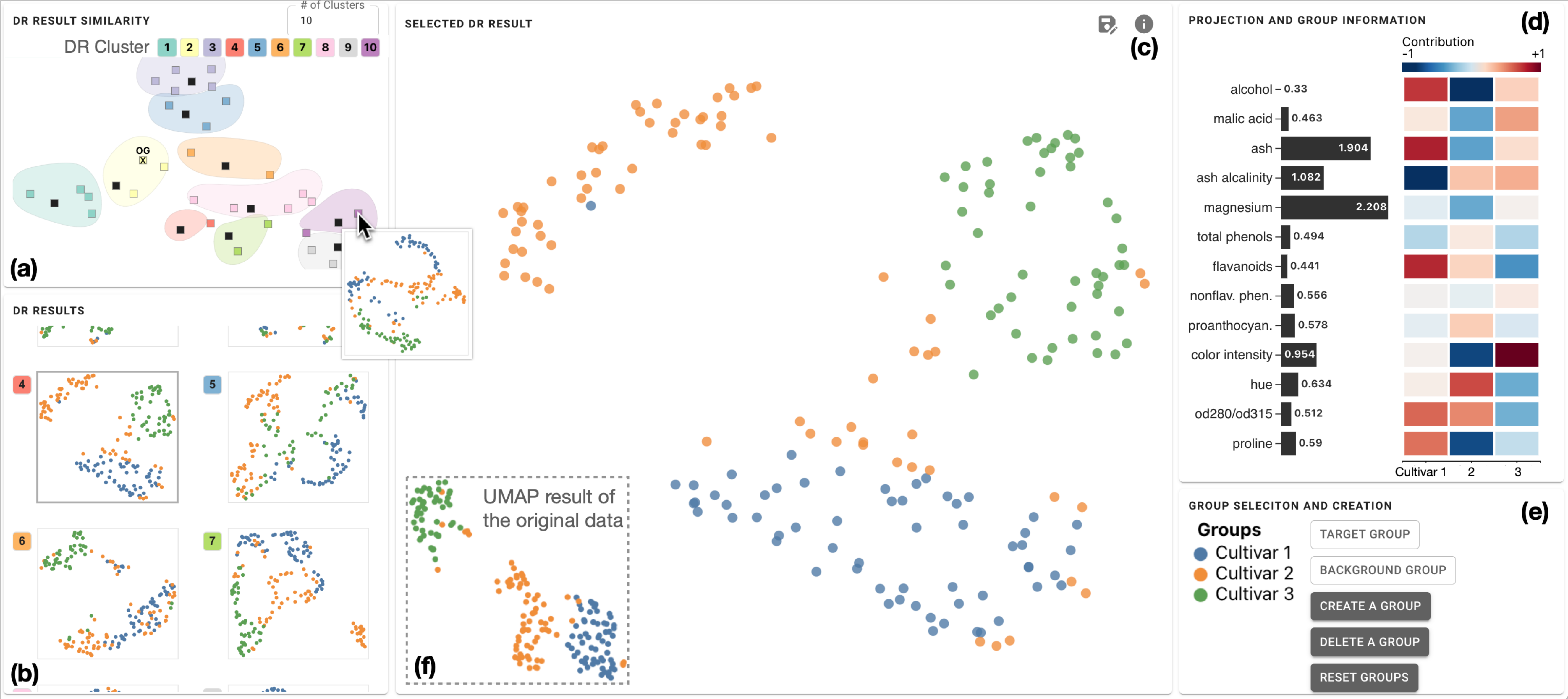}
    \caption{Analyzing the Wine dataset~\cite{uci_mlr} with \methodname{} and the UI. The UI shows the similarities of the generated DR results (a), a representative subset of the DR results (b), a single selected DR result (c), the information for interpreting the DR result (d), and the group information (e). The DR result of the original data is also shown as a reference (f).}
    \label{fig:interface}
\end{figure*}

\noindent\textbf{Quality of the optimization.}
Similar to the above experiment, we generated a single UMAP result with no constraint.
For this experiment, we used the same number of instances ($\nInsts{=}800$) but a different number of attributes ($\nAttrs {=} 5, 10, 20$).
We set $\nLatFeats {=} 2$.
Also, to see the effect of the NMM's settings on the optimization quality, we tested different numbers of evaluations (from 100 to 2000) and initial random solutions ($\nParams {+} 1$, $10 \nParams {+} 1$, $20 \nParams {+} 1)$.
As we do not know the truly best solution, we analyzed relative accuracy to the optimized solution with large numbers of initial solutions and evaluations, specifically, $(50 \nParams {+} 1)$ solutions and 5000 evaluations.
We also set a randomly selected solution as the baseline solution.
Let $v$, $v_\mathrm{best}$, and $v_\mathrm{base}$ denote objective values of \autoref{eq:maximize_graph_dissim} for the comparing, best, and baseline solutions, respectively; then, the relative accuracy is $(v - v_\mathrm{base}) / (v_\mathrm{best} - v_\mathrm{base})$.
As the NMM's solutions can be varied based on the initialization, we computed each of $v$, $v_\mathrm{best}$, and $v_\mathrm{base}$ by taking a mean of 10 trials.

\autoref{fig:comp_eval}-c shows the relative accuracy. 
Generally, the increase in the number of evaluations improves the accuracies. 
Also, our hybrid approach using many initial random solutions improves the accuracies (e.g., ($10p{+}1$) reaches better accuracies than ($p{+}1$)).
Also, we observe that the relative accuracy tends to be higher for the smaller searching space (e.g., the red lines have higher accuracies).
For the $20{\times}2$ searching space, the accuracy relative to $v_\mathrm{best}$ is still low even with 2000 executions and ($20p{+}1$) initial solutions.
Thus, we should use even larger numbers of evaluations and initial solutions to obtain better results.
However, as discussed, the NMM is more suitable for a small search space; thus, PCA or clustering of attributes can be applied for data preprocessing when the original search space is too large.
An analysis example applying PCA can be found in the supplementary materials~\cite{supp}.
For a small search space, our default parameter, $(10p{+}1)$ initial solutions, with 1000 evaluations would provide reasonable results.

\section{Visual Interface}
\label{sec:visual_interface}

To efficiently investigate \methodname{}'s results,  as shown in ~\autoref{fig:interface}, we develop a visual user interface (UI), which is also applicable to other methods developed within \name{}.
The UI is developed as a web application with Python, JavaScript, and D3.
We provide a supplementary demonstration video of the UI~\cite{supp}.

\noindent\textbf{Exploration of DR results.}
The views in \autoref{fig:interface}-a and b are designed for exploration and comparison of the DR results. 
\autoref{fig:interface}-a visualizes the information obtained through
the optimization described in \autoref{sec:optimization}, including the set of UMAP results (i.e., $\smash{\DRResultSet}$), dissimilarities of each result (i.e., $\smash{\DRDissim(\ReprMat_i, \ReprMat_j)}$), and clustering-based recommendations.
To visually convey the dissimilarities of UMAP results, we generate a 2D plot by applying UMAP based on $\smash{\DRDissim(\ReprMat_i, \ReprMat_j)}$ (i.e., UMAP on the UMAP results). 
In this plot, each \textit{square} point corresponds to a single UMAP result and their spatial proximities represent the similarities of the UMAP results.
We indicate each point's belonging cluster by coloring each point and the isocontour generated by Bubble Sets~\cite{collins2009bubble}.
We use black color to distinguish the selected points, which initially correspond to the recommended UMAP results.
Also, a point of the original DR (i.e., $\smash{\ReprMat_0}$) is annotated with the cross mark and text, ``OG''.
We also support fundamental interactions, such as zooming and tooltiping for previewing UMAP results (e.g., one in  \autoref{fig:interface}-a).
A scrollable view in \autoref{fig:interface}-b shows the UMAP results corresponding to $\smash{\ReprMat_0}$ and the selected black points. 
Their belonging clusters are indicated with texts and colored boxes (e.g., 7 with the green box). 
A \textit{circle} point in each UMAP result represents a data instance.
Also, we color these points based on their group labels with a different color scheme from \autoref{fig:interface}-a.
For the comparison, each instance's color is consistent across all the UMAP results. 
Analysts can select one UMAP result from this view for more detailed-level investigations, as explained below.

\noindent\textbf{Interpretation of a DR result.}
To help interpret the selected UMAP result, the view in \autoref{fig:interface}-d shows the information on (1) the projection matrix used to generate the UMAP result and (2) each attribute's contribution to the characteristics of groups in the UMAP result.

The projection matrix, $\smash{\ProjMat}$, contains the information of more (dis)regarded attributes for the generation of the result.
This information is useful to understand the cause of the selected UMAP result's difference from the others.
As described in~\autoref{sec:architecture}, $\smash{\ProjMat}$ can be either diagonal (i.e., $\smash{\ProjMat = \diag(\AttrWeights)}$) or more dense (i.e., when using no constraint or $\smash{\ProjMat = \diag(\AttrWeights) \OrthMan \diag(\LatFeatWeights)}$).
When $\smash{\ProjMat = \diag(\AttrWeights)}$, we visualize values of $\smash{\AttrWeights}$ as a bar chart, as shown on the left side of \autoref{fig:interface}-d.
For the other case, we visualize values in $\smash{\ProjMat}$ as a heatmap using a diverging colormap (e.g., \autoref{fig:case_ccse}-b (left)).

Reviewing patterns shown in the DR result is essential to uncover analytical insights as well as to avoid deriving insights from false patterns due to excessive data transformation.
The patterns are often examined through the comparison of data groups in the DR result~\cite{nonato2019multidimensional,fujiwara2020supporting}.
To assist group comparison, as shown on the right side of \autoref{fig:interface}-d, the UI integrates an existing contrastive-learning-based interpretation method, called ccPCA, and the heatmap-based visualization~\cite{fujiwara2020supporting}.
ccPCA contrasts a target group with a background group to reveal highly-contributed attributes to the characteristics of the target. 
The attributes' contributions are obtained as a weight vector, where the larger magnitude, the stronger contribution to the target group's characteristics.
In addition, the sign of the weight vector can represent the direction of the contribution when using the sign adjustment method~\cite{fujiwara2021visual}.
For example, while \texttt{alcohol} in \autoref{fig:interface}-d  contributes to the characteristics of both \texttt{Cultivars\,1} and \texttt{2},  according to their sign, they likely have higher and lower alcohol percentages than others, respectively. 
Also, to enable the comparison of attribute values of instances, we update the size of each point in \autoref{fig:interface}-c when a certain attribute name is hovered in \autoref{fig:interface}-d.

Although the UI uses predefined labels by default (e.g., the cultivar classes in \autoref{fig:interface}), interactive refinement of groups can be performed with the lasso-selection available in \autoref{fig:interface}-c and controls shown in \autoref{fig:interface}-e.
The changes in groups automatically update the attributes' contributions in \autoref{fig:interface}-d. 
By default, when computing the contributions with ccPCA, each group is selected as a target group and the other groups are set as one background group. 
The UI also allows explicit selection of a background group. 
This is useful when the comparison of two specific groups is more desired.

Through a collective use of the above functionalities and visualizations, we can assess the DR result and patterns. 
When the observed groups do not result from false patterns, the projection matrix values, attributes' contributions, and distribution of 
attribute values should show some consistency.
This is because the separation visible in the DR result should be highly related to the projection matrix, the differences should be captured in the attributes' contributions, and the attributes' contributions should reflect the attribute value distribution. 
We provide a concrete example in our case studies.

\section{Case Studies}

We\,demonstrate the effectiveness of our approach through case studies on real-world datasets. 
Throughout all case studies, we generate UMAP results using two different constraints: $\smash{\ProjMat = \diag(\AttrWeights)}$ and $\smash{\ProjMat = \diag(\AttrWeights) \OrthMan \diag(\LatFeatWeights)}$.
Here we only describe the essential information to present the analysis results.
In the supplementary materials~\cite{supp}, we provide all the other details and one additional case study as an analysis example dealing with a large number of attributes (over 700).
Note that the patterns uncovered in the case studies are difficult to identify with the aforesaid optimization method by Lehmann and Theisel~\cite{lehmann2015optimal} or attribute selection (refer to \cite{supp}). 

\subsection{Study 1: Diverse Categorization of Wines}

We analyze the Wine dataset~\cite{uci_mlr}, which consists of 178 instances, 13 attributes, and cultivar labels.
As shown in \autoref{fig:interface}-f, DR on this dataset usually reveals three clusters highly related to the cultivars.
With \methodname{}, we seek patterns different from those clusters. 

\begin{figure}[t]
    \centering
    \includegraphics[width=\linewidth,height=0.7\linewidth]{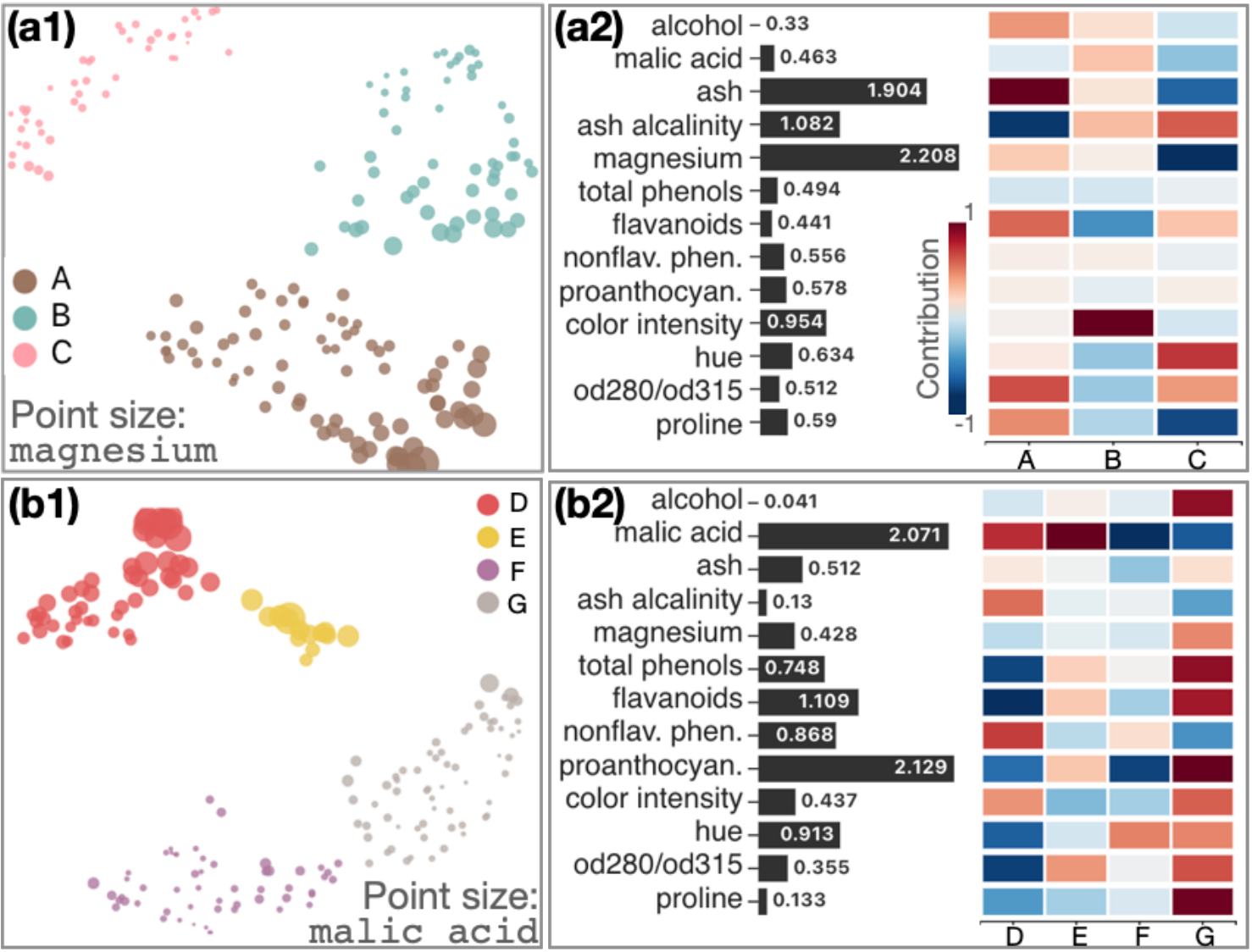}
    \caption{Study 1:\,Diverse categorizations of wines. DR results (a1, b1) selected from the \autoref{fig:interface}-b and their auxiliary information (a2, b2).}
    \label{fig:case_wine}
\end{figure}

As shown in \autoref{fig:interface}-b, \methodname{} produces the results with greatly different patterns. 
We select a UMAP result that contains three clusters (see \autoref{fig:interface}-c).
These clusters (especially, the cluster mainly consisting of \texttt{Cultivar\,1}) seem to contain different wines from the three clusters in the original UMAP result shown in \autoref{fig:interface}-f. 
Also, the proximity of each cluster is clearly different from the original UMAP result (e.g., the clusters mainly consisting of \texttt{Cultivars\,1} and \texttt{3} are placed close with each other in \autoref{fig:interface}-c).
As shown in \autoref{fig:case_wine}-a1, we interactively define the clusters seen in \autoref{fig:interface}-c as \texttt{Groups\,A}--\texttt{C}.
From the auxiliary information displayed in \autoref{fig:case_wine}-a2, the UMAP result is generated with the constraint of $\smash{\ProjMat = \diag(\AttrWeights)}$, where \texttt{ash}, \texttt{ash\,alcalinity}, and \texttt{magnesium} have larger weights than others.
Also, these attributes show strong contributions to each clusters' characteristics, especially for \texttt{Groups\,A} and \texttt{C} (see \autoref{fig:case_wine}-a2(right)).
We further verify the strong associations between the clusters and each of the three attributes. 
For example, as shown in \autoref{fig:case_wine}-a1,  \texttt{Group\,C} has small \texttt{magnesium}. 
According to existing research on this data~\cite{barth2021classification}, only these three attributes represent the mineral content of wines.
Thus, \methodname{} seems to find a new wine categorization that highly corresponds to the mineral content.

\begin{figure*}[t]
    \centering
    \includegraphics[width=\linewidth,height=0.2\linewidth]{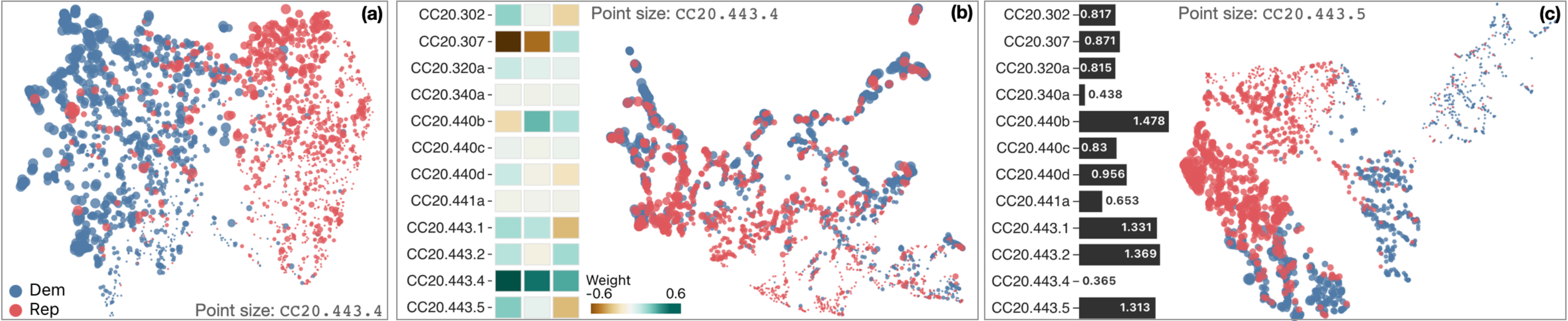}
    \caption{Study 2: Investigation of political opinion patterns in US residents. The original UMAP result (a) and representative \methodname{}'s results produced with different constraints---(b): $\smash{\ProjMat = \diag(\AttrWeights) \OrthMan \diag(\LatFeatWeights)}$ and (c): $\smash{\ProjMat = \diag(\AttrWeights)}$.}
    \label{fig:case_ccse}
\end{figure*}

From \autoref{fig:interface}-b, we select a representative UMAP result of DR Cluster 6 (orange) as another categorization example.
This result contains multiple clusters, each of which is composed of multiple cultivars.
As shown in \autoref{fig:case_wine}-b1 and b2, we create \texttt{Groups\,D}--\texttt{G} and visualize the related auxiliary information.
From the weights and contributions in \autoref{fig:case_wine}-b2, we see that multiple attributes, such as \texttt{malic\,acid}, \texttt{proanthocyanins}, \texttt{flavanoids}, strongly influence the forming of \texttt{Groups\,D}--\texttt{G}. 
These attributes are related to wine taste (e.g., \texttt{proanthocyanins} contributes to the dryness)~\cite{barth2021classification}.
As low weights are assigned for the attributes related to fermentation (i.e., \texttt{alcohol} and \texttt{proline}), mineral content (e.g., \texttt{ash}), and appearance (e.g., \texttt{color\,intensity}), we can say that \methodname{} identifies \texttt{Groups\,D}--\texttt{G} that have the difference more in the taste.

\subsection{Study 2: Investigation of Political Opinion Patterns}

As a case with a larger number of instances, we analyze a survey dataset from the 2020 Cooperative Election Study~\cite{cces_data}, which consists of US residents' responses on various political opinions.
From this dataset, we select 12 ordinal attributes/questions that do not have a high correlation with each other (specifically, less than 0.7 Pearson's correlation coefficient).
We focus on instances/respondents who support either the Democratic (\texttt{Dem}) or Republican (\texttt{Rep}) party and discard instances that have missing values for the 12 attributes.
After the above process, 4462 instances are included in \texttt{Dem} in contrast to 1154 instances in \texttt{Rep}. 
We randomly sampled 1154 instances from \texttt{Dem} to avoid producing patterns related to the sample size difference.
The resulting dataset contains 2308 instances with 12 attributes.
For detailed information of each attribute, refer to \cite{cces_data}.

As in \cite{hare2018ordered}, DR on political survey data of US residents often only reveals two clusters related to the left-right ideology.
In fact, as shown in \autoref{fig:case_ccse}-a, when applying UMAP to the dataset, we can only see such clusters corresponding to \texttt{Dem} and \texttt{Rep}.
We utilize \methodname{} to reveal political opinion patterns hidden in the data. 
Here we describe a few representative findings from our analysis.

As shown in \autoref{fig:case_ccse}-b, we find one UMAP result where \texttt{Dem} and \texttt{Rep} are heavily overlapped with each other.
Because the information of the projection matrix, $\ProjMat$, is shown as a heatmap (see \autoref{fig:case_ccse}-b (left)), the result is generated with the constraint of $\smash{\ProjMat = \diag(\AttrWeights) \OrthMan \diag(\LatFeatWeights)}$.
Based on each attribute weight, we see that \texttt{CC20.307} and \texttt{CC20.443.4} have dominant weights in two of three columns of $\ProjMat$. 
While the question of \texttt{CC20.307} is if the US police make the respondent feel safe or unsafe (1:\,mostly safe--4:\,mostly unsafe), \texttt{CC20.443.4} is how the respondent would like to their legislature to spend money on law enforcement (1:\,greatly increase--4:\,greatly decrease).
As these two attributes' weights have different signs (\texttt{CC20.307}: negative; \texttt{CC20.443.4}: positive), $\ProjMat$ seems to derive new features while debiasing the opinion of future money usage on law enforcement from the current opinion of the police.
We can say that, within the resultant features by $\ProjMat$, there are no clear opinion differences between \texttt{Dem} and \texttt{Rep} unlike the result in \autoref{fig:case_ccse}-a.

\autoref{fig:case_ccse}-c shows another UMAP result, where we can still see the separation between \texttt{Dem} and \texttt{Rep} (\texttt{Dem}'s instances tend to be around the bottom-right side) as well as new clusters that cannot be seen in \autoref{fig:case_ccse}-a.
As shown in \autoref{fig:case_ccse}-c (left), large weights are assigned to \texttt{CC20.440b} (racial problems), \texttt{CC20.443.1}, \texttt{CC20.443.2}, and \texttt{CC20.443.5} (legislature's money use on welfare, healthcare, and transportation/infrastructure, respectively). 
By interactively changing the point size based on each of these attributes, we observe that \texttt{CC20.440b} (racial problems) and \texttt{CC20.443.2} (healthcare) closely associate with the separation between \texttt{Dem} and \texttt{Rep}.
On the other hand, as seen in the sizes of the points in \autoref{fig:case_ccse}-c (right), \texttt{CC20.443.5} is highly related to the clusters aligned along the diagonal direction (e.g., small \texttt{CC20.443.5} can be seen around the top right).
Thus, we can say that the opinions on the money use on transportation/infrastructure are diverse even within each party's supporters. 
By assigning large weights to the above attributes while minimizing the influences from certain attributes (e.g., \texttt{CC20.340a}, political ideology), \methodname{} seems to find political subgroups that are difficult to find with the conventional use of DR.

\vspace{-2pt}
\section{Discussion}
\vspace{-1pt}

\name{} and \methodname{} have provided a primary step in addressing the stated problem of hidden manifolds.
We discuss our approach's limitations as observed through the theoretical and experimental analyses, and discuss further potential enhancements.

\noindent\textbf{Scalability.}
As discussed in \autoref{sec:implementation_and_complexity} and \autoref{sec:computational_eval}, \methodname{} has limited scalability for the numbers of instances ($\nInsts$) and attributes ($\nAttrs$). 
Because of NSD's time complexity, \methodname{} is computationally expensive when $\nInsts$ is large.
Based on the results in \autoref{sec:computational_eval}, \methodname{} is practical up until data with a few thousand instances when using a midrange computer.
When $\nAttrs$ is large, on the other hand, the search space becomes very large; consequently, the NMM requires large numbers of initial solutions and evaluations to find solutions of adequate quality.
Thus, we recommend preprocessing with PCA or attribute clustering for the case with large $\nAttrs$.
One potential approach for efficient optimization is to make all functions differentiable and use derivative-based solvers. 
We expect this can be achieved by utilizing differentiable variants of $k$-neighbor selections~\cite{plotz2018neural} as well as developing NSD for weighted graphs.
While the equations used in ND and NetLSD~\cite{tsitsulin2018netlsd} can be naturally extended weighted graphs, we need further investigations to understand their characteristics in the context of weighted graphs.  

\noindent\textbf{Reliability.} Similar to other ML methods, \methodname{} could suffer from overfitting when $\nInsts$ is relatively small compared to $\nAttrs$~\cite{guo2007regularized}.
Therefore, \methodname{} is suitable for data where $\nInsts$ is considerably larger than $\nAttrs$ (e.g., $n{=}1000$, $m{=}20$).
When $\nInsts$ is relatively small, we can apply PCA to reduce $\nAttrs$ or define stronger constraints on the projection matrix (e.g., only data scaling).
In addition to the projection constraint, \methodname{} has several hyperparameters, $\nLatFeats$, $\LassoCoeff$, $\RidgeCoeff$, $\nEigenvals$, $\NSDParam$, $\nResults$, which can also influence the DR results.
We have suggested default values for $\nEigenvals$ and $\NSDParam$ (i.e., $\nEigenvals{=}50$ and $\NSDParam{=}1$ as explained in \autoref{sec:nsd}), while $\nResults$ (the number of different DR results) can be increased till the convergence of the optimization or set as large as possible based on the available computational power.
The remaining hyperparameters that need to be adjusted are $\nLatFeats$, $\LassoCoeff$, and $\RidgeCoeff$ (the number of latent features in a projection matrix and the weights for L1 and L2 norm-based regularizations).
As in other ML methods, currently, we recommend manually searching appropriate values based on the observed results (e.g., the quality of optimization in \autoref{sec:computational_eval}). 
In future work, we would like to investigate automatic hyperparameter selection.
To reduce the risk of false findings, our visual interface assists the analyst to inspect the obtained DR results with expert knowledge of their data. 

\noindent\textbf{Generalizability.}
\name{} is designed as a general framework for nonlinear DR methods. 
We emphasize that \name{} is applicable to various nonlinear DR methods, such as t-SNE, as discussed in \autoref{sec:exemplifying_method}.
\name{} can also be used for linear DR methods. 
For example, \autoref{eq:maximize_dr_dissim} can be applied to the linear DR method designed by Lehmann and Theisel~\cite{lehmann2015optimal}.
Furthermore, we can extend \autoref{eq:maximize_graph_dissim} to recommend sets of graph-related hyperparameters of DR (e.g., the number of neighbors in UMAP), which produce significantly different DR results~\cite{chatzimparmpas2020tvisne}. 
This can be achieved by replacing a linear projection matrix with hyperparameters for the optimization's search parameters.

\vspace{-1pt}
\section{Conclusion}
\vspace{-1pt}

We have presented \name{}, a feature learning framework that enables investigation of data patterns via a conjoint use with nonlinear DR.
The derived exemplifying method and visual interface have demonstrated the utility of \name{} for analyses of real-world datasets.
This work also exposes the limitations of conventional ways of data exploration using dimensionality reduction and thus contributes toward the maximal utilization of data.

\acknowledgments{
This work has been supported in part by the Knut and Alice Wallenberg Foundation through Grant KAW 2019.0024, the U.S. National Science Foundation through Grant ITE-2134901, and the National Institute of Health through Grant 1R01CA270454-01. 
}

\bibliographystyle{abbrv-doi}

\bibliography{00_ref}
\end{document}